%% file: main.tex
\documentclass[10pt,twocolumn,letterpaper]{article}

\usepackage[pagenumbers]{iccv} 

\input{preamble}

\definecolor{iccvblue}{rgb}{0.21,0.49,0.74}
\usepackage[pagebackref,breaklinks,colorlinks,allcolors=iccvblue]{hyperref}

\def\paperID{10599} 
\def\confName{ICCV}
\def\confYear{2025}

\title{Detail++: Training-Free Detail Enhancer for T2I Diffusion Models}

\author{
Lifeng Chen\textsuperscript{\rm 1}
~~Jiner Wang\textsuperscript{\rm 1}
~~Zihao Pan\textsuperscript{\rm 1}
~~Beier Zhu\textsuperscript{\rm 1, 2}
~~Xiaofeng Yang\textsuperscript{\rm 1, 2}
~~Chi Zhang\textsuperscript{\rm 1}$^{\dagger}$
}

\begin{document}

\twocolumn[{%
\maketitle
\centering
\vspace{-0.7cm}
\noindent
\normalsize\textsuperscript{\rm 1}AGI Lab, Westlake University, \textsuperscript{\rm 2}Nanyang Technological University.

\vspace{0.2cm}
\large\url{https://detail-plus-plus.github.io/}

\begin{center}
    \captionsetup{type=figure}
    \includegraphics[width=\textwidth]{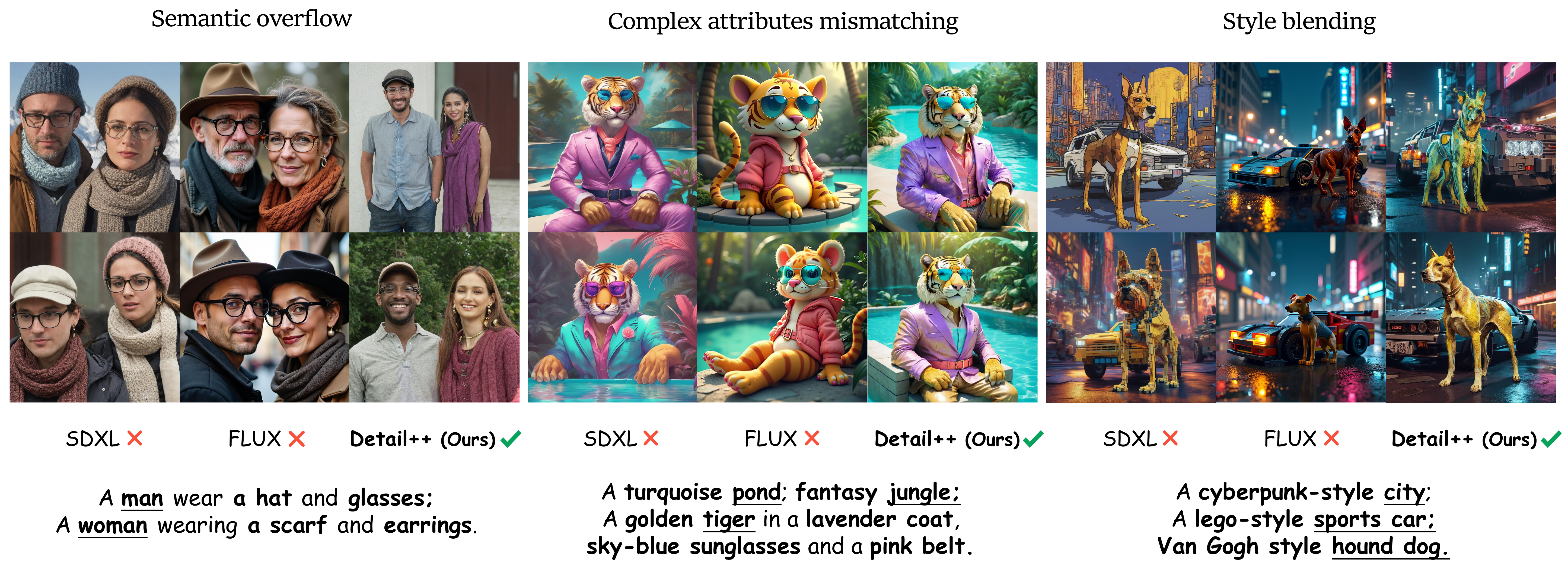}
    \captionof{figure}{\textbf{A comparison between our method and current state-of-the-art generative models.} The mainstream models often suffer from issues such as semantic overflow, complex attribute mismatching, and style blending. Even Flux, the leading generative model under the DiT framework, struggles to overcome these challenges. In contrast, our method, \textbf{Detail++}, based on SDXL, achieves highly accurate semantic binding in a \textbf{\textit{training-free}} way.}
    \label{fig:teaser}
\end{center}
}]

\renewcommand{\thefootnote}{\fnsymbol{footnote}}
\footnotetext[2]{Corresponding author.}
\renewcommand{\thefootnote}{\arabic{footnote}}

\input{sec/0_abstract}    
\input{sec/1_intro}
\input{sec/2_related}
\input{sec/3_preliminary}
\input{sec/4_method}
\input{sec/5_experiment}
{
    \small
    \bibliographystyle{plain}
    \bibliography{main}
}
\input{sec/supplementary}

\end{document}

%% file: preamble.tex
\usepackage{amsmath}
\usepackage{amsfonts}
\usepackage{algorithm}
\usepackage{algpseudocode}
\usepackage{booktabs}
\usepackage{multirow}
\usepackage{array}
\usepackage{graphicx}
\usepackage[table]{xcolor}
\usepackage{pifont}
\usepackage{xspace}
\usepackage{caption}
\usepackage{enumitem}
\usepackage{wrapfig,lipsum}
\usepackage{diagbox}
\usepackage{makecell}
\usepackage{stfloats}
\usepackage{xurl}
\usepackage{tabularx}
\usepackage{fvextra}
\usepackage{subcaption}
\newcommand{\cmarkr}{\textcolor{red}{\ding{51}}\xspace}%
\newcommand{\xmarkr}{\textcolor{red}{\ding{55}}\xspace}%
\newcommand{\xmarkg}{\textcolor{green}{\ding{55}}\xspace}%
\newcommand{\cmarkg}{\textcolor{green}{\ding{51}}\xspace}

\newcommand{\tabincell}[2]{\begin{tabular}{@{}#1@{}}#2\end{tabular}}

\newcommand{\rv}[1]{#1}
\newcommand{\rvtwo}[1]{#1}

\newcommand{\bluenum}[1]{\cellcolor[HTML]{D4E6F1}{\textcolor{black}{#1}}}
\newcommand{\bluetext}[1]{\colorbox[HTML]{D4E6F1}{#1}}
\newcommand{\greentext}[1]{\colorbox[HTML]{E6F8E0}{#1}}
\newcommand{\yellowtext}[1]{\colorbox[HTML]{FFF9C4}{#1}}
\definecolor{softgreen}{HTML}{2E7D32}

%% file: sec/0_abstract.tex
\begin{abstract}
Recent advances in text-to-image (T2I) generation have led to impressive visual results. However, these models still face significant challenges when handling complex prompts—particularly those involving multiple subjects with distinct attributes. Inspired by the human drawing process, which first outlines the composition and then incrementally adds details, we propose Detail++, a training-free framework that introduces a novel Progressive Detail Injection (PDI) strategy to address this limitation. Specifically, we decompose a complex prompt into a sequence of simplified sub-prompts, guiding the generation process in stages. This staged generation leverages the inherent layout-controlling capacity of self-attention to first ensure global composition, followed by precise refinement. To achieve accurate binding between attributes and corresponding subjects, we exploit cross-attention mechanisms and further introduce a Centroid Alignment Loss at test time to reduce binding noise and enhance attribute consistency. Extensive experiments on T2I-CompBench and a newly constructed style composition benchmark demonstrate that Detail++ significantly outperforms existing methods, particularly in scenarios involving multiple objects and complex stylistic conditions.
\end{abstract}

%% file: sec/1_intro.tex
\section{Introduction}
\label{sec:intro}

\begin{figure*}[t]
  \centering
  \includegraphics[width=0.94\linewidth]{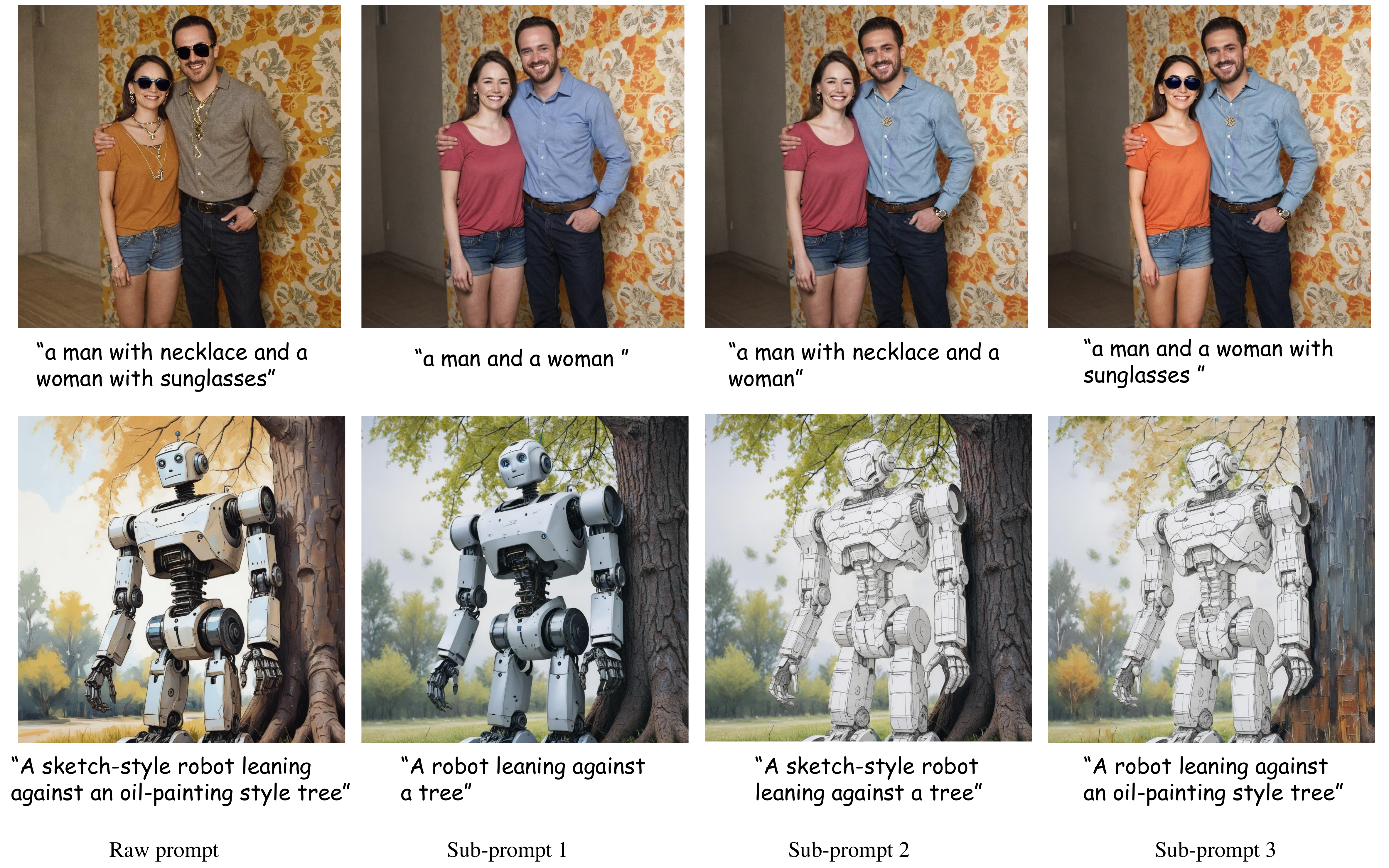}
    \caption{\textbf{The basic process of Detail++.} As shown in the first column, generating complex prompts in a single branch often results in inaccurate or blended attribute assignments. For example, attributes such as ``sunglasses" and ``necklace" may be mistakenly applied to the wrong subject. Our method addresses this challenge through a progressive approach: we first ignore all complex modifiers to produce a rough generation base, then systematically inject details to ensure each attribute is precisely added to its corresponding subject region. The prompt displayed below each image indicates the input used for that specific branch, and the second row demonstrates the method's effectiveness in style combination scenarios. Note that all four branches here are generated in parallel.}
  \label{fig:progressive}
\end{figure*}

In recent years, text-to-image (T2I) generation~\cite{saharia2022photorealistic,rombach2022high,esser2024scaling,podell2023sdxl,flux2024} techniques have advanced remarkably, enabling the creation of high-quality, detail-rich images from textual descriptions and demonstrating considerable potential in art design, advertising, entertainment, and education~\cite{lei2024stylestudio,li2024unbounded,dehouche2023s}. Despite these improvements, current methods still struggle when handling complex descriptions that involve multiple subjects. A well-known challenge, termed ``detail binding,'' emerges when descriptions assign distinct sets of attributes to multiple subjects. This often causes models to misassociate attributes across subjects, resulting in semantic overflow, incorrect attribute matching, and undesired style blending, as shown in Fig. ~\ref{fig:teaser}.

Regarding the above challenges, our observation is that current state-of-the-art models attempt to render all elements and their attributes from a prompt simultaneously, leading to imperfect detail binding. Drawing inspiration from human artistic practice—where artists typically begin by sketching the basic spatial layout and outlines before gradually refining details—we want to progressively inject attributes to accurately integrate specific details into their corresponding subject regions, thereby achieving more semantically coherent image generation. To achieve our goal, we propose \textbf{Detail++}, a multi-branch image generation framework where different simplified versions of the prompt work together to produce accurate image content. The process begins with an initial denoising branch using the original prompt, during which we extract and share self-attention maps across other following branches to maintain consistent layout. 
The second denoising branch employs the simplest sub-prompt—only containing just the subject and its basic layout—to create a foundation without fine details. In subsequent parallel branches, specific attributes from each sub-prompt are progressively injected into the image, with all branches maintaining the same overall layout. To ensure proper binding of details to their correct subjects, we introduce an Accumulative Latent Modification strategy that uses cross-attention maps to generate binary masks, precisely locating different subject regions for targeted detail injection. 
Additionally, we incorporate Centroid Alignment Loss during test time to optimize subject-specific cross-attention maps, addressing map inaccuracies and further improving detail injection precision. The whole process is illustrated in Fig.~\ref{fig:progressive}.

Our proposed Detail++ was evaluated on the widely used T2I-CompBench benchmark~\cite{huang2023t2i} as well as on our custom style composition benchmark. The results demonstrate that Detail++ consistently outperforms existing methods, particularly in scenarios involving multiple style bindings. In qualitative evaluations, \textbf{Detail++} produces high-quality images without additional fine-tuning~\cite{hu2024ella,jiang2024comat,huang2023t2i,feng2024ranni} or predefined layout information~\cite{li2023gligen,zheng2023layoutdiffusion,feng2024ranni,lian2023llm,zhang2024realcompo,zhao2023loco}, underscoring the practicality of our approach. 
The main contributions of this paper are as follows:
\begin{itemize}
    \item We innovatively propose \textbf{Detail++}, a training-free multi-branch framework. It improves the accuracy of existing T2I models when facing complex prompts involving multiple subjects through progressive attribute injection.
    
    \item We developed a self-attention map sharing mechanism that utilizes the properties of self-attention map in U-Net to control the image layout, thus creating a consistent base for accurate attribute binding.
    
    \item We further introduced a test-time optimization strategy based on Centroid Alignment Loss, which further improved the accuracy of attribute binding.

    \item Extensive experiments demonstrate the advantages of Detail++ compared to current SOTA methods, and we will subsequently open-source the code.
\end{itemize}

%% file: sec/2_related.tex
\section{Related work}
\label{sec:related}
\subsection{Text-to-Image Diffusion Models}
\label{sec2.1}
Diffusion Models have become the dominant approach in text-to-image (T2I) generation~\cite{balaji2022ediff,podell2023sdxl,ramesh2021zero,rombach2022high,dhariwal2021diffusion,flux2024,esser2024scaling}. Models like Stable Diffusion~\cite{rombach2022high}, DALL-E 2~\cite{ramesh2022hierarchical}, and Imagen~\cite{saharia2022photorealistic} generate high-quality images by progressively denoising noise. They leverage neural networks to map text to visuals, often using pre-trained text encoders like CLIP~\cite{radford2021learning} to enhance text comprehension and guide generation. However, even the most advanced models, such as Flux~\cite{flux2024}, still struggle to generate images that closely align with the text prompts. This has led to a line of studies focused on improving T2I alignment.

\subsection{Text-to-Image Alignment}
As mentioned earlier, T2I baseline models struggle to precisely follow text conditions, often facing issues such as attribute overflow (where attributes spill over to unmentioned subjects), mismatching (where attributes are incorrectly matched), and blending (where attributes from different subjects mix onto a single subject). To address these challenges, several solutions have been proposed~\cite{wu2024core,wei2024enhancing,kim2024text,ge2023expressive,chen2024cat,feng2022training,park2023energy}. Among them, some methods based on test-time optimization~\cite{zhang2024enhancing,zhang2024object,li2023divide,chefer2023attend,rassin2023linguistic,agarwal2023star,park2023energy,cho2023visual,zhang-wan-2025-r} aim to align the cross-attention maps of attributes with the subject tokens, but they are often slow and difficult to optimize for noise efficiently. Other methods based on layout~\cite{zhang2024realcompo,lian2023llm,feng2024ranni,zhang2023controllable,yang2024mastering,qu2023layoutllm} rely on users or LLMs to predefine the layout, often involving complex intermediate steps. There are also approaches that need fine-tuning~\cite{hu2024ella,chen2023geodiffusion,jiang2024comat,yang2023reco,feng2024ranni,zhangitercomp,lv2025rethinking}, such as ELLA~\cite{hu2024ella}, which enhances the model's understanding of text by integrating LLMs into T2I generation models, but requires significant modifications to the baseline model, leading to high training costs. While these methods have made notable progress in addressing attribute mismatching, they still struggle with overflow and blending. \rv{Recently, a unified-model reinforcement learning approach, T2I-R1\cite{jiang2025t2i}, has been proposed, which enables the model to deliberate over the textual prompt before autoregressively generating image tokens, demonstrating substantial potential for improving alignment in text-to-image generation.} However, the attribute blending issue remains largely unresolved. In contrast, our method, Detail++, effectively mitigates these challenges by progressively injecting attributes, thereby enabling accurate binding across diverse attribute types, including object, color, texture, and style.

\subsection{Image Editing}
Image editing manipulates specific regions while preserving overall fidelity. Traditional text-driven image editing methods~\cite{li2020manigan,nam2018text,xia2021tedigan} have shown promise by combining GANs~\cite{gal2022stylegan,patashnik2021styleclip} with CLIP. Recently, a large number of new methods~\cite{balaji2022ediff,kim2022diffusionclip,brooks2023instructpix2pix,meng2021sdedit,xie2023smartbrush} based on diffusion models have emerged to address various aspects of image editing tasks in a training-free manner, most of which~\cite{tumanyan2023plug,nam2024dreammatcher,hertz2022prompt,liu2024towards,cao2023masactrl} leverage the attention mechanism~\cite{vaswani2017attention}. For example, Prompt-to-Prompt (P2P)~\cite{hertz2022prompt} adjusts cross-attention maps to enable a variety of editing operations. MasaCtrl~\cite{cao2023masactrl} realizes rigid editing while maintaining the overall textures and identity in a mutual attention way. FPE~\cite{liu2024towards} distinguishes the different functionalities between self-attention and cross-attention in stable diffusion models. \rv{Furthermore, \cite{shin2025exploring} extends this research into DiT-based model, and finds similar features with U-Net-based models.} Our method draws inspiration from these works and further extends them to a progressive editing framework.

%% file: sec/3_preliminary.tex
\section{Preliminary}
\label{sec:preliminary}

This section introduces the basics of the attention mechanism, which serves as a key component of our proposed method.

\begin{figure}[t]
  \centering
  \includegraphics[width=1\linewidth]{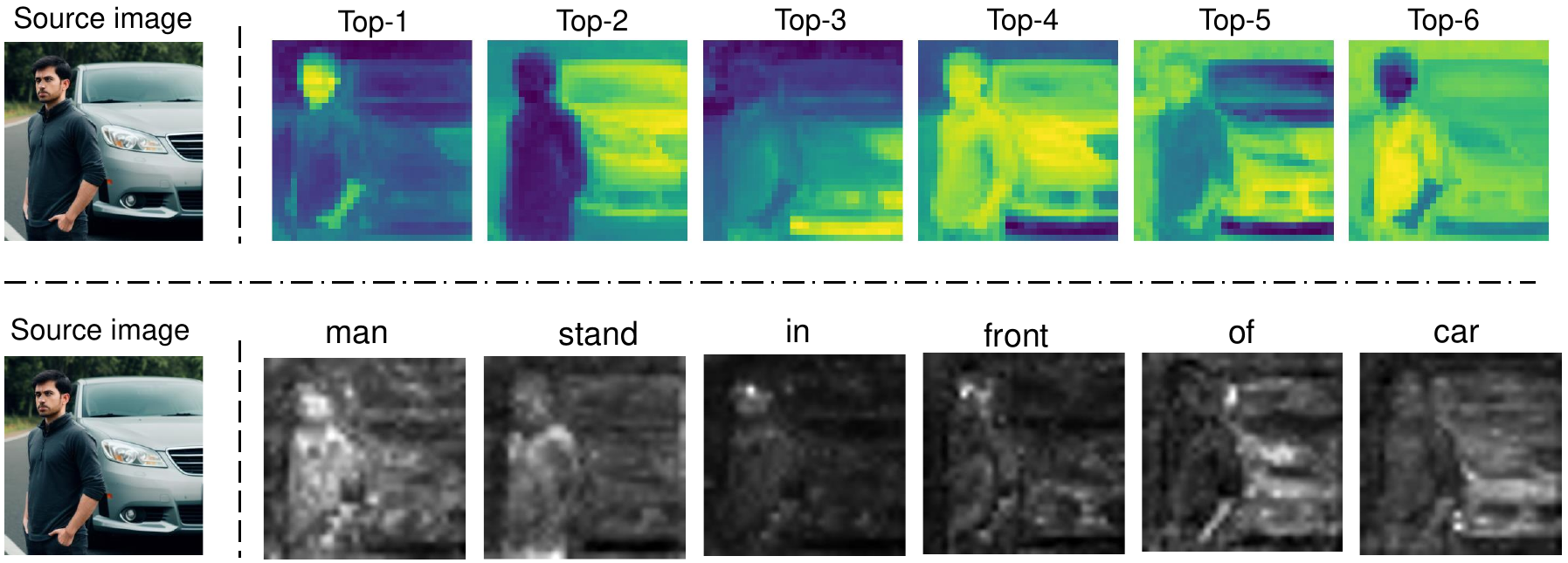}
  \caption{\textbf{Attention visualization.} The visualization of the self-attention map and cross cross-attention map of the prompt ``\textit{man stand in front of car}". The self-attention map is visualized by displaying the top-6 components obtained after SVD~\cite{wall2003singular}. The cross-attention map visualizations correspond to each token in the prompt.
  }
  \label{fig:comparison}
\end{figure}

\textbf{Attentions in U-Net.}\label{pre:attn_in_unet} The U-Net architecture~\cite{ronneberger2015u} has become a cornerstone in many state-of-the-art diffusion.
U-Net in diffusion models integrates both self-attention and cross-attention mechanisms, each serving unique purposes in image generation. 
Cross-attention focuses on extracting the semantic elements of the prompt, ensuring the generated images are consistent with text prompt~\cite{hertz2022prompt}. The attention matrix in cross-attention can be defined as:
\begin{equation} \label{eq:cross_attention}
M_{\text{cross}} = \text{Softmax}\left(\frac{Q_{\text{img}} K_{\text{text}}^\top}{\sqrt{d_k}}\right),
\tag{1}
\end{equation}
where $Q_{\text{img}}$ is the query embedding derived from the U-Net's internal features, $K_{\text{text}}$ is the key embedding originating from the textual embeddings of the prompt, and $d_k$ is the dimensionality of the key vectors. 
On the other hand, self-attention primarily relates to the spatial layout and structural details of the image by computing relationships within the image feature space~\cite{liu2024towards}. This attention map can be represented as:
\begin{equation} \label{eq:self_attention}
M_{\text{self}} = \text{Softmax}\left(\frac{Q_{\text{img}} K_{\text{img}}^\top}{\sqrt{d_k}}\right),
\tag{2}
\end{equation}
where both $Q_{\text{img}}$ and $K_{\text{img}}$ are the query and key matrices derived from the internal features within the U-Net. The self-attention map provides an internal relational structure that is crucial for preserving spatial consistency. To better understand the difference between self-attention and cross-attention, we visualize the attention maps of them separately in Fig.~\ref{fig:comparison}.
Our proposed method builds upon these foundational attention mechanisms, specifically modifying and extending them to improve the handling of complex prompts. 

%% file: sec/4_method.tex
\makeatletter
\newcommand{\rmnum}[1]{\romannumeral #1}
\newcommand{\Rmnum}[1]{\expandafter\@slowromancap\romannumeral #1@}
\makeatother

\begin{figure*}[t]
  \centering
  \includegraphics[width=1\linewidth]{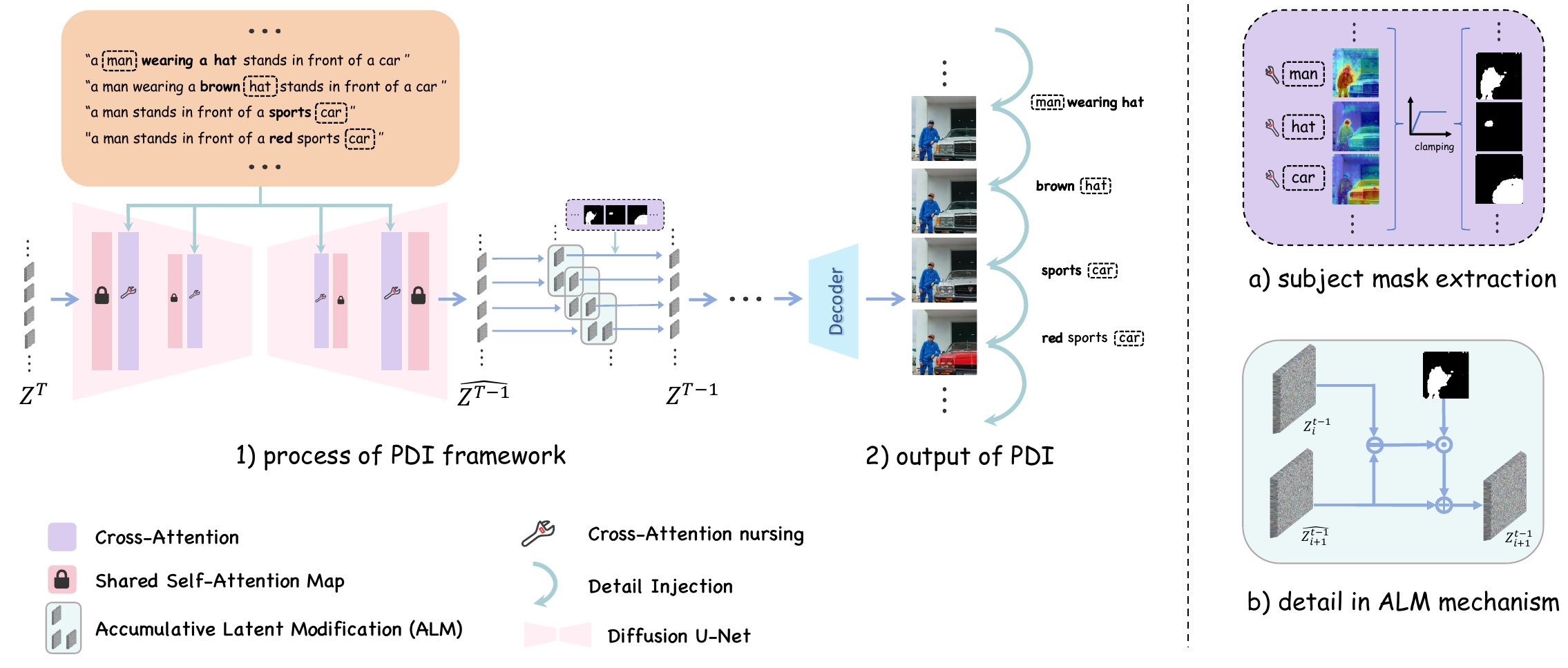}
 \caption{\textbf{Overview of Detail++.} 1) Firstly, all sub-prompts are passed to U-net for generation at the same time. In each denoising step, we first share the SA map of the first branch with all other branches to ensure a consistent layout base. Then, for each timestep generated latent batch, we (a) extract each subject's mask from its token's cross-attention map, and (b) use these masks in our Accumulative Latent Modification to inject the new attributes only into the corresponding subject regions. Finally, after denoising all time steps, we can obtain 2) detailed images whose attributes are progressively added. In addition, to obtain each more accurate subject mask, we perform test-time optimization on the corresponding subject's cross-attention map based on our proposed centroid alignment loss.
  }
  \label{fig:pipeline}
\end{figure*}

\section{Method}
\label{sec:method}
\subsection{Overview}
In this section, we provide an overview of our method, \textbf{Detail++}. Our core idea is to first ignore all complex, hard-to-generate details to obtain a rough generation base, and then progressively add more details in a painting-like manner. Given a complex original prompt, we automatically generate a series of simplified prompts, as described in Sec.~\ref{sec:prompt segment}. 
To ensure that images generated from these prompts with different attributes maintain the same layout, we share the self-attention map obtained during the generation of the original prompt with all sub-prompt generation processes (Sec.~\ref{sec:unified attention}). This shared self-attention map provides a stable structural foundation.
Controlled by our Accumulative Latent Modification strategy(Sec.~\ref{sec:Progressive latent editing.}), correct details are progressively injected into the corresponding subject regions. At each timestep, cross-attention maps for each subject keyword are used to create precise masks that guide the selective attribute injection. 
Moreover, in Sec.~\ref{sec:attention nursing}, we introduce a Centroid Alignment Loss to achieve more accurate and focused cross-attention map during test time, ensuring that the added details are strictly confined to their intended subject regions. The overall workflow is illustrated in Fig.~\ref{fig:pipeline}.

\subsection{Progressive Detail Injection}
\label{sec:PDI framewor}
\textbf{Prompt decomposition.}\label{sec:prompt segment} To handle the complexity of detailed prompts, the first step in our method is to simplify the input prompt. Specifically, we use a language model, such as spaCy~\cite{honnibal2017spacy} or ChatGPT~\cite{brown2020languagemodelsfewshotlearners} to decompose the original complex prompt \( p_0 \) into a sequence of progressively simplified sub-prompts, denoted as \( \mathcal{P} = \{ p_0, p_1,\) \(\dots, p_{n-1}, p_n \} \), where \( p_1 \) represents the most simplified version by removing all modifiers. This simplification ensures that all attributes are eliminated in the \( p_1 \) branch, providing a clear base for progressive detail injection. Sub-prompts \( p_2, p_3,\) \(\dots, p_n \) each add an additional modifier to \( p_1 \). The choice of decomposition is explained in more detail in Supplementary Sec.~I-C.

Furthermore, for each pair of prompt \( p_{i+1} (i>0) \) and \( p_1 \), we also use the language model to identify the subject \( q_i \) to which the newly added attribute in \( p_{i+1} \) corresponds. These subjects are represented by the set \( \mathcal{Q} = \{ q_1,\dots,q_{n-2},q_{n-1}\} \).
For example, given the prompt \( p_{0} =  \)\textit{``a red teddy bear wearing a green tracksuit''}, the set \( \mathcal{P} \) and subject set \( \mathcal{Q} \) would be as follows:

\[
\mathcal{P} = \left\{
\begin{array}{l}
    p_0:\textit{``a red teddy bear wearing a green tracksuit''}, \\
    p_1:\textit{``a teddy bear wearing a tracksuit''}, \\
    p_2:\textit{``a red teddy bear wearing a tracksuit''}, \\
    p_3:\textit{``a teddy bear wearing a green tracksuit''} 
\end{array} \right\},
\]
\[
\mathcal{Q} = \left\{
\begin{array}{l}
    q_1:\textit{``teddy bear''}, \\
    q_2:\textit{``tracksuit''} 
\end{array} \right\}.
\]

\textbf{Shared self-attention map.}\label{sec:unified attention} Once we have the set of sub-prompts, the next step is to use these prompts to generate images cooperatively. Since previous work~\cite{liu2024towards,tumanyan2023plug,cao2023masactrl,nam2024dreammatcher} has shown that the self-attention map in U-Net stores the layout information of the image, we let the network generate images based on each sub-prompt, while achieving a consistent layout by sharing the self-attention map. Specifically, at each denosing time step, we first cache the self-attention maps in U-Net during the first branch generation process and reuse them in other branches. In this way, we make the images that different sub-prompts generated consistent with the layout of the original prompt. It is observed that the early stages of denoising are most critical for constructing the overall layout of the image, so we only use this strategy during the initial \( S \) steps of the denoising process. For the remaining \( T-S\) steps, we allow each sub-prompt's U-Net model to predict noise independently to ensure the image fidelity, as is done in the standard diffusion process. In this way, we can keep the layout of the images generated by the different sub-prompts consistent with the original prompt.

\textbf{Accumulative Latent Modification.}\label{sec:Progressive latent editing.} In order to precisely bind the newly added attributes in each sub-prompt to their corresponding subjects, we further develop latent-level subject masking for each attribute injection. Specifically, we extract a binary mask based on the cross-attention map corresponding to each subject \( q_i \)~(Sec.~\ref{sec:prompt segment}) from a sub-prompt, which highlights the relevant regions of each subject in the image. This process involves normalizing the cross-attention maps and applying a thresholding operation to create a distinct binary mask for each subject \( q_i \). The binary mask extraction for subject \( q_i \) is given by the following formula:

\begin{equation}
\label{eq:mask_compute}
B_{i} = \mathbf{1}\left[\frac{\overline{M_{i}} - \min(\overline{M_{i}})}{\max(\overline{M_{i}}) - \min(\overline{M_{i}})} > \tau\right],
\tag{3}
\end{equation}
where $B_{i}$ is the binary mask of subject $q_i$, and $\overline{M_{i}}$ represents its averaged attention map across layers (Supplementary Sec.~I-G). $\tau$ is a threshold parameter. The indicator function $\mathbf{1}[\cdot]$ converts the normalized attention values to a binary mask, where values above $\tau$ are set to 1 and others to 0. $\min(\overline{M_{i}})$ and $\max(\overline{M_{i}})$ represent the minimum and maximum values of the averaged attention map respectively.

Binary masks can restrict latent modifications to a specific region. Specifically, for the $i$-th branch, the modified latent representation is computed as:
\begin{equation}
\label{eq:progressive_injection}
z_{i+1}^{t-1} = z_i^{t-1} + B_i \odot (\widehat{z_{i+1}^{t-1}} - z_i^{t-1}),
\tag{4}
\end{equation}
where $\widehat{z_{i+1}^{t-1}}$ represents the denoised latent in $i+1$-th branch, $z_i^{t-1}$ is the latent from the previous branch, $B_i$ is the binary semantic mask for the current subject, and $z_{i+1}^{t-1}$ is the modified latent incorporating correct attribute. The operator $\odot$ denotes element-wise multiplication.

For example, if a region's mask value $b$ is 1, this signifies that the region corresponds to the target subject. In this case, the new latent $z_{i+1}$ generated with the additional prompt detail is applied to this region, ensuring our framework integrates the added details precisely within the designated area. Conversely, if $b$ is 0 for a given region, indicating that the area is unrelated to the current attribute, the model reuses the latent $z_{i}$ from the previous branch, thereby excluding the new attribute's influence from unrelated regions. This selective application prevents cross-subject interference and preserves the distinction of each subject's features across the image. Thus, the entire process of progressive attention substitution is outlined in Algorithm~\ref{alg:peudo}.
\begin{algorithm}
\caption{Progressive Detail Injection}
\label{alg:peudo}
\begin{algorithmic}[1]
\Require A source prompt $p_{0}$, and sub-prompts derived from it ${p_1, ..., p_{n}}, \text{DM}$ represents the diffusion models to predict the noise of next step. $T$ is the total denoising timesteps.
\Ensure Input $\mathcal{P}=\{p_0, p_1, \dots, p_n\}$ to model in parallel.
\State $z_0^T, z_1^T, \dots, z_n^T \gets z^T \sim \mathcal{N}(0, \mathcal{I})$; \Comment{Initialized with same latent}.
\State $\mathcal{Z}^T\gets\{z_0^T, z_1^T, \dots, z_n^T\}$
\For{$t = T, \dots, T-S+1$}
    \State $\mathcal{Z}^{t-1} \gets \text{DM}(\mathcal{Z}^t, \mathcal{P}, t,\overline{M_{self}^{t}})$;\Comment{$\overline{M_{self}^{t}}$ is from first branch}.
    \For{$i = 1, 2, \dots, n-1$}
        \State $\widehat{z_{i+1}^{t-1}} \gets z_{i+1}^{t-1}$
        \State $z_{i+1}^{t-1} \gets z_{i}^{t-1} + B_i \odot (\widehat{z_{i+1}^{t-1}} - z_{i}^{t-1})$;\Comment{$B_i$ is the bianry mask consistent with Sec 4.2.3.}
    \EndFor
\EndFor
\For{$t=T-S, \dots, 1$}
    \State $\mathcal{Z}^{t-1} = \text{DM}(\mathcal{Z}^{t},\mathcal{P},t)$;
\EndFor
\State \Return $\mathcal{Z}^0$
\end{algorithmic}
\end{algorithm}

\subsection{Centroid Alignment Loss}
\label{sec:attention nursing}
When generating binary masks from cross-attention maps, we observe a critical issue in existing cross-attention mechanisms: attention activations tend to be dispersed and unfocused, causing a subject's attention to spread into other subject regions, resulting in attribute injection errors and degraded generation quality.
To address this problem, we propose a test-time optimization method by introducing a Centroid Alignment Loss to refine cross-attention maps. Our key insight is that in ideal conditions, a subject's cross-attention map should form concentrated activations at the center of the subject region, rather than being scattered across the entire image space. Based on this, we design a mechanism that encourages each subject's sailient attention point (the point with largest attention value) to align with the centroid of its attention distribution.

Specifically, we first calculate the centroid $p_{\text{centroid}}(q_i)$ for each subject $q_i$'s attention map, where $q_i$ represents the subject word to which the attribute in the $i$-th branch is being added. This centroid serves as the weighted center of the cross-attention map corresponding to $q_i$'s token in the prompt:
\begin{equation}
\label{eq}
p_{\text{centroid}}(q_i) = \frac{1}{\sum_{h,w} \overline{M_i}(h, w)} 
\begin{bmatrix} 
\sum_{h,w} w \cdot \overline{M_i}(h, w) \\ 
\sum_{h,w} h \cdot \overline{M_i}(h, w) 
\end{bmatrix},
\tag{6}
\end{equation}
where $h$ and $w$ represent the height and width spatial coordinates respectively, and \( \overline{M_i}(h, w) \) is the normalized attention value at that position.

Next, we define the Centroid Alignment Loss by minimizing the Euclidean distance between the attention centroid and the brightest point $p_{\text{max}}(q_i)$, encouraging a more focused attention distribution:
\begin{equation}
\label{eq:alignment loss}
    L_{\text{align}} = \sum_i \left\| p_{\text{centroid}}(q_i) - p_{\text{max}}(q_i) \right\|^2.
\tag{7}
\end{equation}
We combine this loss term with the entropy loss used in ToME\cite{hu2025token} to form the total loss function $L_{\text{total}}$. The entropy loss is defined as:
\begin{equation}
L_\text{ent} = -\sum\nolimits_{m \in \overline{M_i}} m \log(m),
\tag{8}
\end{equation}
where $m$ represents the attention values in the normalized cross-attention map $\overline{M_i}$ for subject $q_i$. Our total loss function becomes $L_{\text{total}} = L_{\text{align}} + \lambda L_\text{ent}$, where $\lambda$ is a trade-off hyperparameter. During each diffusion step \( t \), we update the latent variable through gradient descent:
\begin{equation}
\label{eq:latent update}
    \mathbf{z}'_{t} \leftarrow \mathbf{z}_t - \alpha_t \cdot \nabla_{\mathbf{z}_t} \, L_{\text{total}},
\tag{9}
\end{equation}
where \( \alpha_t \) is the step size.

Experimental results demonstrate that introducing the Centroid Alignment Loss not only produces more concentrated cross-attention maps but also significantly improves the precision of attribute injection. The optimized attention maps generate more accurate binary masks, ensuring that detailed attributes are correctly injected into their corresponding subject regions, effectively resolving attribute confusion issues in complex prompts.

\subsection{\rv{Extension to DiT-based Backbones}}
\label{sec:dit implm}
\rv{A major advantage of Detail++ is its plug-and-play nature. Given that most recent text-to-image diffusion models \cite{esser2024scaling,flux2024,wu2025qwen} are built upon Diffusion Transformers (DiT), whose architecture differs substantially from U-Net based designs, we further extend Detail++ to DiT-based implementations. In particular, we integrate our method into the multi-modal attention (MM-Attention) blocks that are characteristic of DiT-style text-to-image models.}

\rv{Let $\mathbf{X}^{\text{txt}} \in \mathbb{R}^{L_t \times d}$ and $\mathbf{X}^{\text{img}} \in \mathbb{R}^{L_i \times d}$ denote the text and image tokens respectively. In MM-Attention, each modality first computes its own query, key, and value tensors via modality-specific linear projections, e.g., $\mathbf{Q}^{\text{txt}} = \mathbf{X}^{\text{txt}} \mathbf{W}_Q^{\text{txt}}$ and $\mathbf{Q}^{\text{img}} = \mathbf{X}^{\text{img}} \mathbf{W}_Q^{\text{img}}$, with analogous definitions for $\mathbf{K}^{\text{txt}}, \mathbf{V}^{\text{txt}}$ and $\mathbf{K}^{\text{img}}, \mathbf{V}^{\text{img}}$. The text and image branches are then concatenated along the token dimension, i.e., $\mathbf{Q} = [\mathbf{Q}^{\text{txt}}; \mathbf{Q}^{\text{img}}]$, $\mathbf{K} = [\mathbf{K}^{\text{txt}}; \mathbf{K}^{\text{img}}]$, and $\mathbf{V} = [\mathbf{V}^{\text{txt}}; \mathbf{V}^{\text{img}}]$, and a single head's attention map is computed as
\begin{equation}
\mathbf{M} = \operatorname{softmax}\!\left(\frac{\mathbf{Q}\mathbf{K}^\top}{\sqrt{d_k}}\right)
\in \mathbb{R}^{(L_t + L_i) \times (L_t + L_i)}
\tag{10}
\label{eq:mm_attn}
\end{equation}
Because we place all text tokens before image tokens, the attention map $\mathbf{M}$ in Eq.~\eqref{eq:mm_attn} can be partitioned into four blocks,
\begin{equation}
\mathbf{M} =
\begin{bmatrix}
\mathbf{M}_{\text{T}\rightarrow\text{T}} & \mathbf{M}_{\text{T}\rightarrow\text{I}} \\
\mathbf{M}_{\text{I}\rightarrow\text{T}} & \mathbf{M}_{\text{I}\rightarrow\text{I}}
\end{bmatrix},
\tag{11}
\label{eq:mm_block}
\end{equation}
where the bottom-right block encodes image-to-image interactions and the bottom-left block encodes how image queries attend to text keys. We therefore explicitly define
\begin{equation}
\mathbf{M}^{\text{I2I}} = \mathbf{M}_{\text{I}\rightarrow\text{I}},
\qquad
\mathbf{M}^{\text{T2I}} = \mathbf{M}_{\text{T}\rightarrow\text{I}},
\tag{12}
\label{eq:i2i_t2i}
\end{equation}
which serve as the I2I and T2I attention maps that will be used by Detail++ in the DiT backbone.}

\rv{Analogous to the U-Net case, we treat $\mathbf{M}^{\text{I2I}}$ and $\mathbf{M}^{\text{T2I}}$ in Eq.~\eqref{eq:i2i_t2i} as the counterparts of the self-attention and cross-attention maps used in the U-Net-based backbone. During sampling, $\mathbf{M}^{\text{I2I}}$ is propagated to subsequent DiT blocks to enforce a consistent spatial layout through the denoising process, thereby stabilizing the coarse structure of the generated image. In parallel, we derive binary mask from $\mathbf{M}^{\text{T2I}}$ for each subjects(e.g., by aggregating and thresholding the text-to-image scores over text tokens), and use this mask to progressively refine fine-grained attributes in a prohressive manner.}

\begin{figure}[t]
  \centering
  \includegraphics[width=0.98\linewidth]{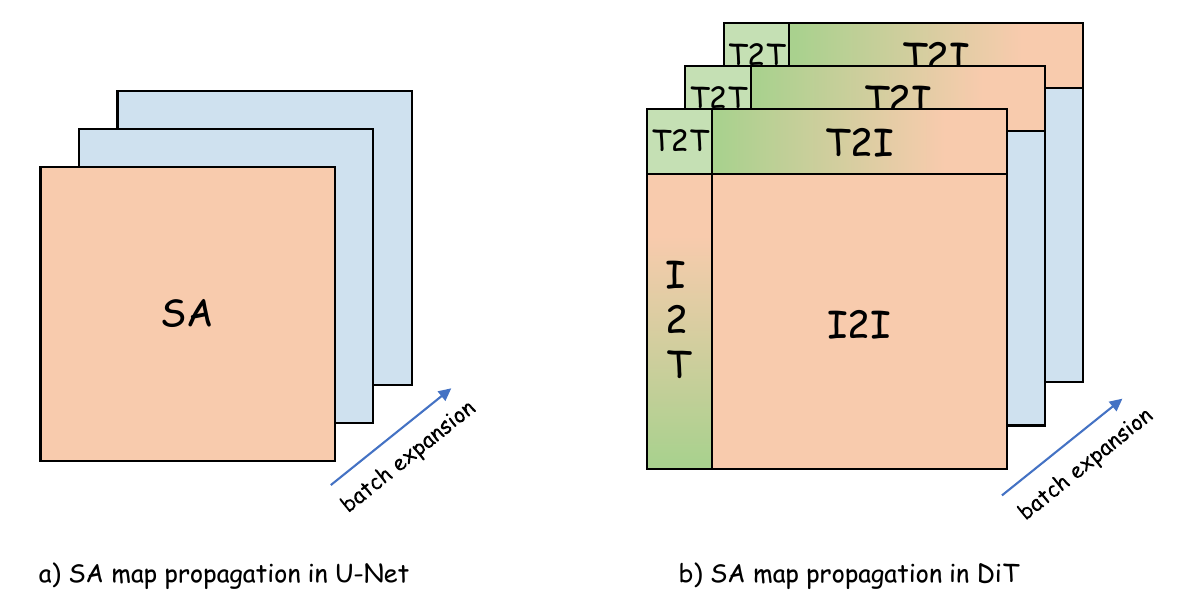}
  \caption{\rv{Efficient self-attention propagation for multi-branch sampling. The self-attention map from the first branch is expanded along the batch dimension and reused by all subsequent branches, avoiding redundant self-attention computation in (a) U\mbox{-}Net and (b) DiT backbones.}}
  \label{fig:sa-prop}
\end{figure}

\subsection{Efficient Self-Attention Propagation}
\label{sec:effcient SA}
To further reduce the computational overhead of multi-branch inference, we propose an Efficient Self-Attention Propagation scheme, as illustrated in Fig.~\ref{fig:sa-prop}. The core idea is to compute the full self-attention weight matrix only once in the first branch \rvtwo{and cache it as $\mathbf{M}$. For each subsequent branch $i$, we skip the attention computation and only compute its branch-specific value projection $\mathbf{V}_i$. The cached $\mathbf{M}$ then directly replaces the branch-wise self-attention weights in both U\mbox{-}Net and DiT backbones}, yielding the output
\begin{equation}
\mathbf{O}_i = \mathbf{M}\mathbf{V}_i.
\tag{13}
\end{equation}
\rvtwo{This reformulation is mathematically equivalent to the original shared-map pipeline in Sec.~\ref{sec:unified attention} and introduces no approximation.} Since in both U\mbox{-}Net- and DiT-based diffusion models the sequence length of image tokens is typically much larger than that of text embeddings, this strategy effectively reduces the memory consumption of multi-branch sampling while preserving semantic controllability and content diversity across branches.

In addition, we introduce a branch scheduler to further improve inference efficiency. Specifically, during the final $T\!-\!S$ denoising steps, where self-attention map propagation is no longer required, we retain only the last branch and discard the remaining ones. \rvtwo{The last branch is conditioned on the full prompt, ensuring that the complete semantic description of all subjects is preserved throughout the remainder of the denoising process.} This design substantially reduces both memory usage and runtime overhead during inference. A detailed comparison of efficiency gains is provided in our ablation study (see Table~\ref{tab:eff_ablation}).

%% file: sec/5_experiment.tex
\begin{table*}[h]
\caption{Quantitative comparison of detail binding performance across multiple metrics. The highest scores are highlighted in \bluetext{blue}, second-highest in \greentext{green}, and third-highest in \yellowtext{yellow}.}
\label{table:quantitative results}
\centering
\resizebox{\textwidth}{!}{%
\begin{tabular}{lc  |ccc  |ccc  |c}
\toprule
\multirow{2}{*}{Method} & \multirow{2}{*}{Train} &
\multicolumn{3}{c|}{BLIP-VQA $\uparrow$} &
\multicolumn{3}{c|}{Human-preference $\uparrow$} &
\multirow{2}{*}{SCB $\uparrow$} \\ 
& & Color & Texture & Shape & Color & Texture & Shape & \\ 
\midrule
SD1.5\cite{rombach2022high} & \cmarkr
& 0.4719 & 0.4334 & 0.3898
& -2.278 & -1.115 & -1.741 & 0.2162 \\

Composable Diffusion\cite{liu2022compositional} & \xmarkg
& 0.4063 & 0.3645 & 0.3299
& -2.009 & 0.747 & -0.819 & 0.2024 \\

Structured Diffusion\cite{feng2022training} & \xmarkg
& 0.4990 & 0.4900 & 0.4218
& 0.074 & 0.912 & -0.124 & 0.2195 \\

GORS\cite{huang2023t2i} & \cmarkr
& 0.6603 & 0.6287 & 0.4785
& 0.239 & 1.752 & -0.105 & 0.2168 \\

Attention Regulation\cite{zhang2024enhancing} & \xmarkg
& 0.5860 & 0.5173 & 0.4672
& 0.201 & 1.454 & -0.794 & 0.2223 \\

ELLA(SD1.5)\cite{hu2024ella} & \cmarkr
& 0.6911 & 0.6308 & 0.4938
& 0.400 & 1.730 & 0.016 & 0.2366 \\

SDXL\cite{podell2023sdxl} & \cmarkr
& 0.6369 & 0.5637 & 0.5408
& 0.503 & 1.663 & -0.075 & 0.2467 \\

Iter-Comp\cite{zhangitercomp} & \cmarkr
& 0.7114 & 0.6517 & 0.5447
& 0.746 & 1.829 & 0.714 & 0.2534 \\

Ranni(SDXL)\cite{feng2024ranni} & \cmarkr
& 0.6893 & 0.6325 & 0.4934
& - & - & - & - \\

ToMe\cite{hu2025token} & \xmarkg
& 0.6583 & 0.6371 & 0.5517
& 0.424 & 1.564 & 0.563 & 0.2525 \\

PixArt-$\alpha$\cite{chen2023pixart} & \cmarkr
& 0.6886 & 0.7044 & 0.5582
& -0.584 & 1.709 & 1.020 & 0.2368 \\

SD3\cite{esser2024scaling} & \cmarkr
& 0.8017 & 0.7650 & 0.6003
& 0.675 & 1.843 & 0.037 & 0.2503 \\

Flux.1.dev\cite{flux2024} & \cmarkr
& 0.7424 & 0.6446 & 0.5767
& 0.791 & 1.875 & 1.189 & 0.2511 \\

RPG\cite{yang2024mastering} & \xmarkg
& 0.7400 & 0.6547 & 0.5873
& 0.684 & 1.861 & 1.620 & 0.2579 \\

R-Bind(SD3)\cite{zhang-wan-2025-r} & \xmarkg
& 0.8143 & 0.7736 & 0.6021
& 0.667 & 1.899 & 1.274 & 0.2588 \\

T2I-R1\cite{jiang2025t2i} & \cmarkr
& 0.8432 & 0.7653 & 0.6123
& 0.874 & 1.922 & 1.538 & 0.2598 \\

TACA(SD3)\cite{lv2025rethinking} & \cmarkr
& 0.8027 & 0.7504 & 0.6024
& 0.781 & 1.863 & 1.260 & 0.2512 \\

TACA(Flux)\cite{lv2025rethinking} & \cmarkr
& 0.7752 & 0.6584 & 0.5171
& 0.821 & 1.908 & 1.650 & 0.2507 \\

\midrule
Detail++(SDXL) & \xmarkg
& 0.7389 & 0.7241 & 0.5582
& \greentext{0.934} & \greentext{1.922} & \yellowtext{1.842}
& \yellowtext{0.2624} \\

Detail++(SD3) & \xmarkg
& \bluetext{0.8703} & \bluetext{0.8699} & \greentext{0.6100}
& \yellowtext{0.925} & \bluetext{1.965} & \greentext{1.851}
& \greentext{0.2689} \\

Detail++(Flux) & \xmarkg
& \greentext{0.8461} & \greentext{0.8348} & \yellowtext{0.6051}
& \bluetext{1.084} & \yellowtext{1.933} & \bluetext{1.876}
& \bluetext{0.2696} \\

\bottomrule
\end{tabular}
}
\end{table*}

\begin{figure*}[tp]
    \centering
    \includegraphics[width=0.99\linewidth]{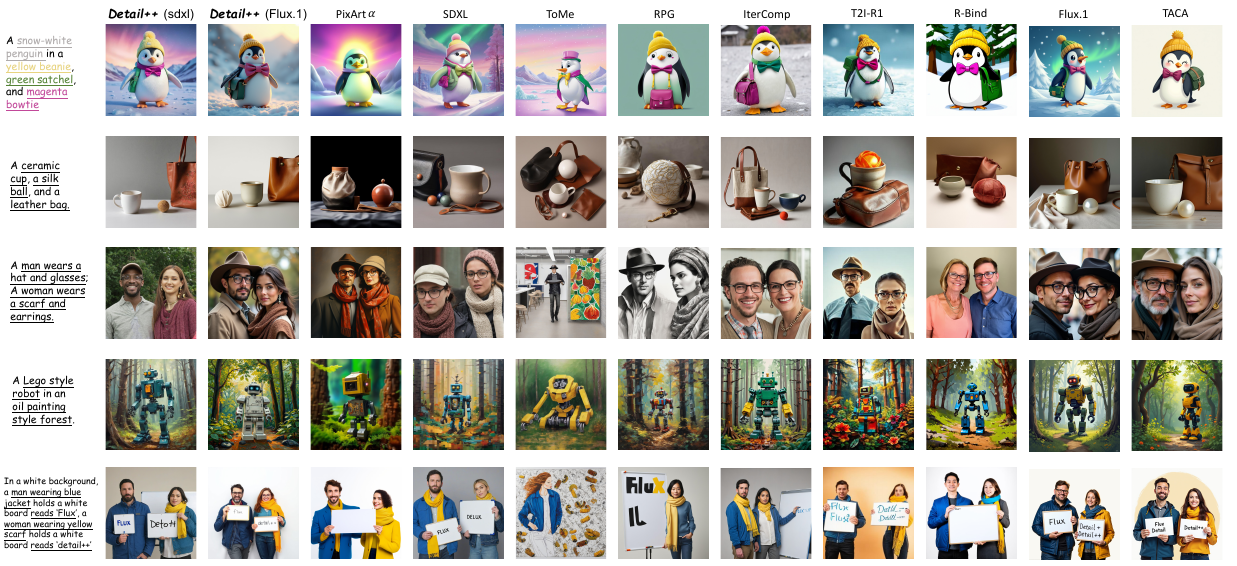}
    \caption{Qualitative comparison of various methods on complex prompts with multiple attribute types (object, color, texture, style \rv{and text}). Our method effectively prevents attribute overflow, complex attribute mismatching, and style blending while maintaining high visual fidelity.}
    \label{fig4:qualitative_comparison}
\end{figure*}

\begin{figure*}[tp]
    \centering
    \includegraphics[width=0.99\linewidth]{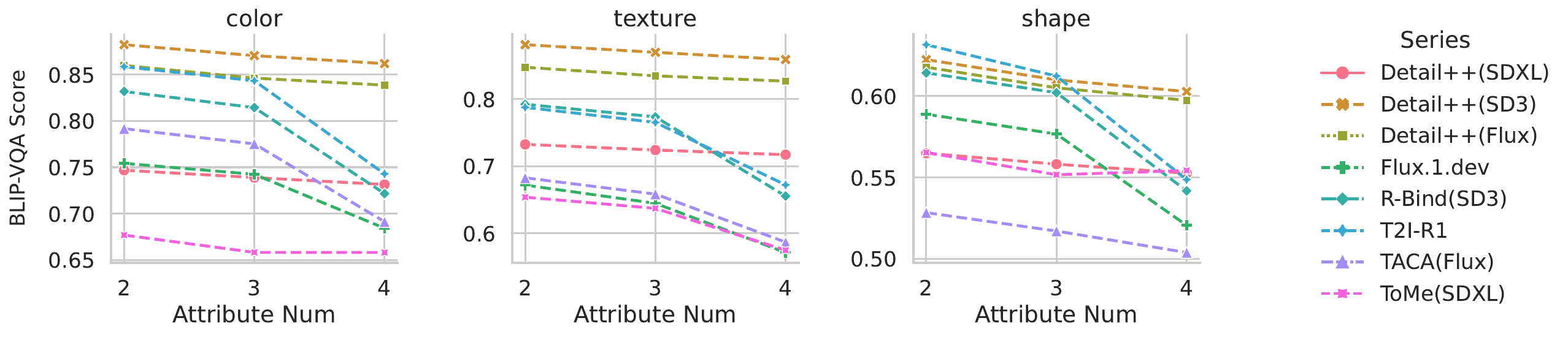}
    \caption{\rv{Quantitative comparison of baselines under different numbers of attributes. Detail++ (Flux) and Detail++ (SD3) exhibit near-stable performance as the attribute count increases, consistently ranking among the top three methods. Detail++ (SDXL) also enters the top-three group when the number of attributes increases to four.}}
    \label{fig:attribute_performance}
\end{figure*}

\begin{table*}[t]
\caption{\rvtwo{Relative VRAM and inference time overhead (\%) of each method compared to its baseline with fused SDPA enabled, measured under prompts with 3, 4, and 5 attributes (Attr3/4/5).} Higher overhead is highlighted in \textcolor{red}{red}. Lower overhead is highlighted in \textcolor{softgreen}{green}.}
\label{tab: efficiency increase}
\centering
\footnotesize
\setlength{\tabcolsep}{5pt}
\renewcommand{\arraystretch}{1.08}

\makebox[\textwidth][c]{%
\resizebox{0.88\textwidth}{!}{%
\begin{tabular}{l l ccc ccc}
\toprule
\multirow{2}{*}{\textbf{Method}} &
\multirow{2}{*}{\textbf{Baseline}} &
\multicolumn{3}{c}{\textbf{VRAM Overhead (\%)}} &
\multicolumn{3}{c}{\textbf{Time Overhead (\%)}} \\
\cmidrule(lr){3-5}\cmidrule(lr){6-8}
& & \textbf{Attr3} & \textbf{Attr4} & \textbf{Attr5}
  & \textbf{Attr3} & \textbf{Attr4} & \textbf{Attr5} \\
\midrule
Detail++ & SDXL &
\textcolor{softgreen}{+32.2\%} & \textcolor{softgreen}{+49.4\%} & \textcolor{softgreen}{+59.9\%} &
\textcolor{softgreen}{+48.1\%} & \textcolor{softgreen}{+92.6\%} & \textcolor{softgreen}{+122.2\%} \\
ToMe & SDXL &
\textcolor{red}{+115.2\%} & \textcolor{red}{+115.2\%} & \textcolor{red}{+115.3\%} &
\textcolor{red}{+418.5\%} & \textcolor{red}{+551.9\%} & \textcolor{red}{+655.6\%} \\
Detail++ & SD3 &
\textcolor{softgreen}{+6.9\%} & \textcolor{softgreen}{+8.7\%} & \textcolor{softgreen}{+12.5\%} &
\textcolor{softgreen}{+46.5\%} & \textcolor{softgreen}{+52.4\%} & \textcolor{softgreen}{+58.7\%} \\
R-Bind & SD3 &
\textcolor{red}{+78.1\%} & \textcolor{red}{+78.1\%} & \textcolor{red}{+78.1\%} &
\textcolor{red}{+52.4\%} & \textcolor{red}{+58.7\%} & \textcolor{red}{+77.8\%} \\
Detail++ & Flux &
+8.8\% & +10.9\% & +11.9\% &
\textcolor{softgreen}{-83.0\%} & \textcolor{softgreen}{-78.7\%} & \textcolor{softgreen}{-77.0\%} \\
\bottomrule
\end{tabular}%
}}
\end{table*}

\begin{figure*}[tp]
    \centering
    \includegraphics[width=0.94\linewidth]{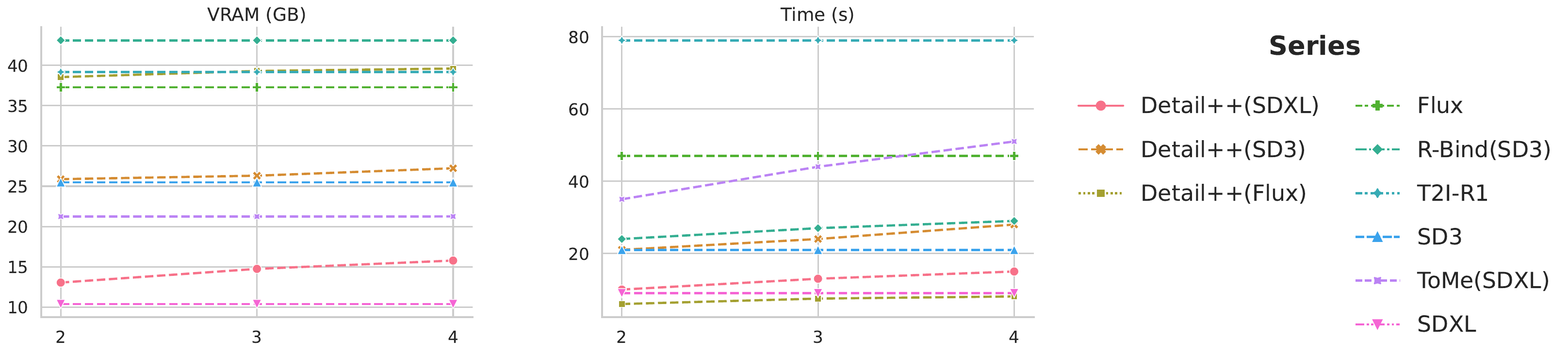}
    \caption{\rv{Comparison of inference time and average VRAM usage across different methods. With the efficient self-attention propagation strategy, increasing the number of branches in Detail++ incurs only a modest overhead in both inference latency and memory consumption.}}
    \label{fig:efficiency_comparison}
\end{figure*}

\section{Experiments}
\label{sec:expr}
\subsection{Experimental Setups}
\textbf{Evaluation Benchmarks and Metrics.}We conduct extensive experiments on the T2I-CompBench++\cite{huang2023t2i}, a widely-used benchmark designed specifically for text-to-image generation tasks involving complex compositional prompts. Specifically, we adopt subsets in T2I-CompBench++ designed to evaluate attributes such as color, shape, texture, because they directly reflect our method’s capacity for semantic alignment. To further validate the subjective quality of generated images and assess human preference scores, we incorporate the ImageReward\cite{xu2023imagereward} model, a learned metric widely accepted for reflecting human perceptual judgments in image generation tasks.
 
Existing benchmarks typically lack an effective evaluation mechanism for attributes related to the combination of multiple artistic styles, which facing the problems of style blend. To bridge this gap and provide a more comprehensive evaluation, we propose a novel benchmark called the Style Composition Benchmark (SCB). \rv{SCB consists of 1000 carefully designed prompts that cover a wide range of distinguished artistic styles binding with multiple subjects.} More details can be found in Supplementary Sec.~I-D.

Our evaluation method for SCB independently measures the alignment between each element in the image and its corresponding style descriptor. Specifically, it involves parsing the prompt of each generated image to identify style descriptors associated with different semantic components. Each component in the generated image is first cropped out using Grounding DINO~\cite{liu2024grounding}, based on the component words parsed from the prompt. We then calculate the component-style correspondence for each segmented element using the CLIP-Score~\cite{radford2021learning,hertz2024style} model. The final style score for each generated image is computed as the average of the style-matching scores across all segmented semantic elements. Therefore, based on the deviation from the reference images, our new metric can efficiently assess whether style blending has occurred.

\begin{figure*}[t]
  \centering
  \includegraphics[width=0.93\linewidth]{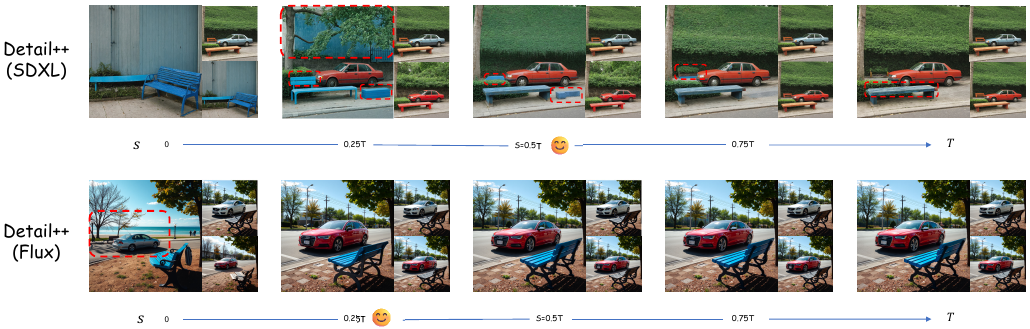}
  \caption{\rv{Qualitative ablation on the self-attention map propagation step $S$ under different backbone architectures.
  \rvtwo{The prompt is ``a blue bench and a red car''; each group shows branch~0 (top-right), branch~1 (bottom-right), and the final branch (large left image).}}
  \rvtwo{Red bounding boxes mark suboptimal generation for the final branch compared with branch~0.
  As $S$ increases, as the red box highlighed, these inconsistencies progressively diminish.
  At $S\!=\!T$, texture degradation may appear in attribute-modified regions (e.g., the blue bench surface highlighted in Detail++(SDXL)'s output), as the layout constraint spans the entire denoising process and leaves no steps for local detail refinement.}}
  \label{fig:ablation-n-self}
\end{figure*}

\begin{figure}[t]
  \centering
  \includegraphics[width=0.98\linewidth]{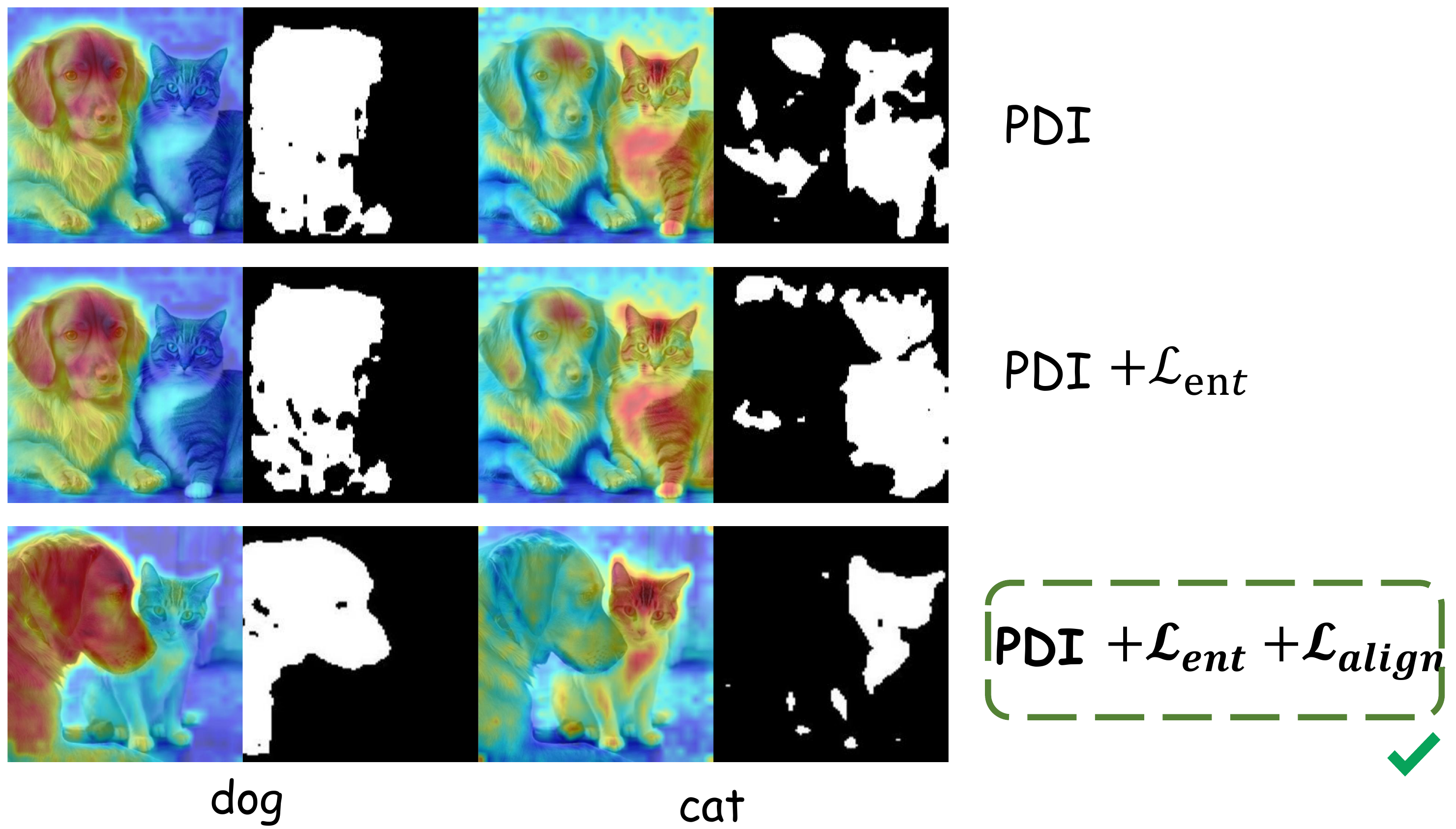}
  \caption{Ablation study of different optimization terms in binary mask extraction. The combined use of both $L_{ent}$ and $L_{align}$ produces more accurate and focused subject masks compared to using only $L_{ent}$ or no optimization.}
  \label{fig:ablation-ca-map}
\end{figure}

\begin{figure}[t]
  \centering
  \includegraphics[width=0.98\linewidth]{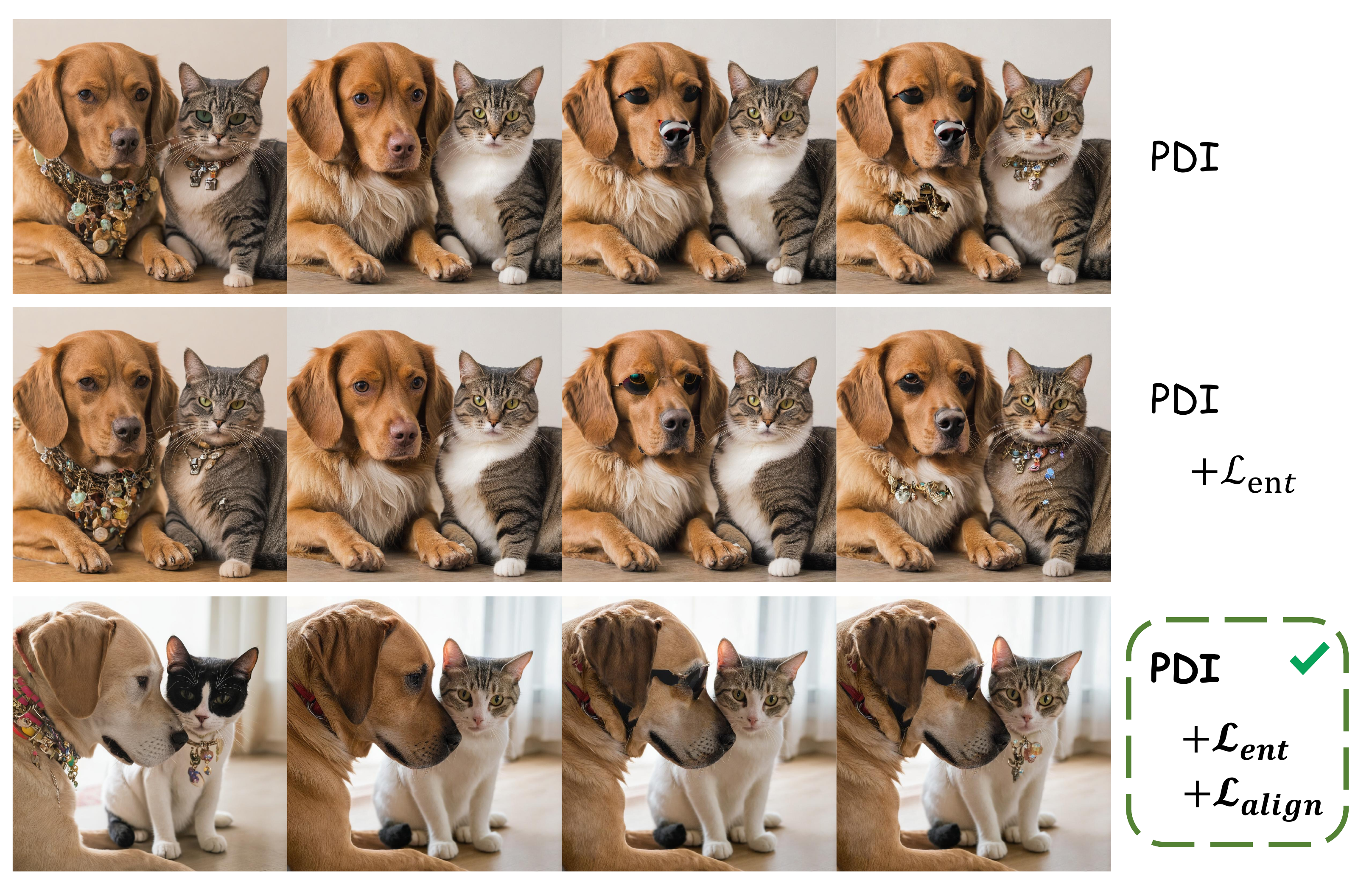}
  \caption{Visual results of different optimization strategies for the prompt ``a dog wearing sunglasses and a cat wearing a necklace''. All results in 3 rows are generated based on the same random seed. The combined use of both $L_{ent}$ and $L_{align}$ achieves more precise detail injection, preventing attributes from incorrectly spilling into adjacent subject regions.}

  \label{fig:ablation-results}
\end{figure}

\noindent\textbf{Implementation Details.}
\rv{We build our implementation upon SDXL~\cite{podell2023sdxl}, SD3 Medium~\cite{esser2024scaling}, and Flux.1-schnell~\cite{flux2024} baselines. Since the attribute-binding prompts in T2ICompBench++~\cite{huang2023t2i} typically follow the pattern ``A [Attribute A] [Subject A] and [Attribute B] [Subject B]'', we use the spaCy~\cite{honnibal2017spacy} NLP toolkit to automatically identify modifier/attribute spans and manage prompt parsing in a reproducible manner. For efficiency, we only retain and test the generation from the last branch.}

\rv{For SDXL, the self-attention map sharing mechanism is activated during the first 50\% of the total denoising steps ($S = 0.5T$), where we share self-attention maps of spatial size $32 \times 32$ from all blocks. For both SD3 and Flux, the sharing mechanism is activated during the first 25\% of the total denoising steps ($S = 0.25T$), using self-attention maps of spatial size $64 \times 64$ from top-5 with highest activations blocks.} For test-time optimization, we set $\lambda = 1$ to achieve a good balance between the two losses. Additional experimental and implementation details are provided in Supplementary Sec.~I-B.

\noindent\textbf{Comparison Methods.} To comprehensively evaluate the performance of our proposed approach, we conduct comparative analyses against several methods. \rv{In addition to the established baseline methods, including SDXL~\cite{podell2023sdxl} and Flux~\cite{flux2024}, we compare against a diverse set of approaches explicitly designed to enhance fine-grained detail binding and compositionality. These include methods that leverage external large language models (LLMs) to provide region-level guidance~\cite{yang2024mastering,feng2024ranni}, approaches based on test-time optimization~\cite{zhang2024enhancing,hu2025token,zhang-wan-2025-r}, fine-tuning based solutions~\cite{hu2024ella,zhangitercomp,lv2025rethinking}, as well as reinforcement learning (RL) based fine-tuning strategies built upon unified multimodal models~\cite{jiang2025t2i}.}

\begin{table}[t]
  \caption{Ablation study of different optimization terms on T2I-CompBench++. The combined use of both $L_\text{ent}$ and $L_\text{align}$ performs best.}
  \centering
  \small  
  \setlength{\tabcolsep}{10pt}  
    \begin{tabular}{ccc|ccc}
    \toprule
    \multirow{2}{*}{\tabincell{c}{$PDI$}} & \multirow{2}{*}{$\mathcal{L}_\text{ent}$} & \multirow{2}{*}{$\mathcal{L}_\text{align}$} & \multicolumn{3}{c}{BLIP-VQA$\uparrow$} \\ 
    & & & Color   & Texture   & Shape  \\ 
    \midrule
     \xmarkr     & \xmarkr & \xmarkr &0.6369&0.5637&0.5408\\
     \cmarkg     & \xmarkr & \xmarkr &0.6942   &0.6836&0.5507\\
     \cmarkg     & \cmarkg & \xmarkr &0.7034& 0.6995&0.5521\\
     \cmarkg     & \cmarkg & \cmarkg &\textbf{0.7389}&\textbf{0.7241}&\textbf{0.5582}\\ 
    \bottomrule
    \end{tabular}
  \label{tab:ablation}
\end{table}

\begin{table}[t]
\centering
\caption{\rv{Efficiency ablation of our inference optimizations.
Batch expansion reuses the first-branch self-attention map by expanding it along the batch dimension, while the branch scheduler retains only the last branch in late denoising steps.
\rvtwo{The reported BLIP-VQA score is the average over the color, texture, and shape subsets on the T2I-CompBench++ test set.}}}
\label{tab:eff_ablation}

\setlength{\tabcolsep}{9pt}
\renewcommand{\arraystretch}{1.15}
\resizebox{\columnwidth}{!}{
\begin{tabular}{@{}l l l l@{}}
\toprule
\textbf{Setting} & \textbf{VRAM $\downarrow$} & \textbf{Time $\downarrow$} & \textbf{BLIP-VQA $\uparrow$} \\
\midrule
Detail++(SDXL) & 19.2GB & 17s & 0.6737\phantom{$_{\textcolor{red}{\scriptscriptstyle\downarrow 1.42\%}}$} \\
+ batch expansion &
16.61GB$_{\textcolor{softgreen}{\scriptscriptstyle\downarrow 13.54\%}}$ &
16s$_{\textcolor{softgreen}{\scriptscriptstyle\downarrow 5.88\%}}$ &
0.6734$_{\textcolor{red}{\scriptscriptstyle\downarrow 0.045\%}}$ \\
+ branch scheduler &
13.05GB$_{\textcolor{softgreen}{\scriptscriptstyle\downarrow 32.03\%}}$ &
10s$_{\textcolor{softgreen}{\scriptscriptstyle\downarrow 41.18\%}}$ &
0.6641$_{\textcolor{red}{\scriptscriptstyle\downarrow 1.42\%}}$ \\
\midrule
Detail++(Flux) & 46.85GB & 23s & 0.7620\phantom{$_{\textcolor{red}{\scriptscriptstyle\downarrow 1.33\%}}$} \\
+ batch expansion &
43.10GB$_{\textcolor{softgreen}{\scriptscriptstyle\downarrow 8.00\%}}$ &
21s$_{\textcolor{softgreen}{\scriptscriptstyle\downarrow 8.70\%}}$ &
0.7617$_{\textcolor{red}{\scriptscriptstyle\downarrow 0.039\%}}$ \\
+ branch scheduler &
37.51GB$_{\textcolor{softgreen}{\scriptscriptstyle\downarrow 19.94\%}}$ &
9s$_{\textcolor{softgreen}{\scriptscriptstyle\downarrow 60.87\%}}$ &
0.7519$_{\textcolor{red}{\scriptscriptstyle\downarrow 1.33\%}}$ \\
\bottomrule
\end{tabular}
}
\end{table}


\subsection{Experimental Results}
\textbf{Quantitative Comparison.}
As shown in Table~\ref{table:quantitative results}, our method achieves superior performance in terms of color, texture, and is comparable on shape binding on T2I-CompBench++~\cite{huang2023t2i}, as well as improved style binding on our proposed SCB benchmark, when compared to existing methods. Furthermore, using ImageReward~\cite{xu2023imagereward} as a proxy metric for human preference, Detail++ demonstrates stronger alignment with human judgments.

\rv{Importantly, our approach can be applied in a plug-and-play manner to both U\mbox{-}Net- and DiT-based diffusion models, consistently yielding stable improvements over their respective baselines across public benchmarks. Specifically, Detail++ improves the overall binding performance by $15.90\%$ (relative) on SDXL, $11.10\%$ (relative) on SD3, and $16.31\%$ (relative) on Flux, highlighting its robustness and architectural generality.}

\noindent\textbf{Qualitative Comparison.}
Fig.~\ref{fig4:qualitative_comparison} presents a qualitative comparison illustrating how different methods handle four representative prompt categories: complex color binding, complex texture binding, object binding, and style composition. \rv{To further demonstrate the effectiveness of Detail++ on DiT-based backbones, we additionally include a text-rendering case.} From the first two rows, it can be observed that when the number of subjects increases, attribute mismatching for color and texture becomes more pronounced. \rv{Even for DiT-based models, noticeable color mixing still occurs, and the visual fidelity and realism of the generated images tend to degrade as the models attempt to accommodate highly complex prompts.} In the third row, which focuses on object-type attributes, semantic leakage frequently arises, causing subjects without explicit modifiers to be unintentionally influenced by attributes assigned to other objects. In the fourth row, we observe that existing mainstream methods struggle to disentangle multiple style modifiers applied to different components, often resulting in severe style blending. Our method, however, is the only one that successfully generates a Lego-style robot while preserving a pure oil-painting background, \rv{leading to a visually clear and strong contrast between foreground and background.} \rv{In the last row, even advanced models may exhibit interference between text rendering and the semantic meaning of the rendered words; for example, rendering the word ``Detail++'' can introduce repetitive or ambiguous artifacts.} Overall, the qualitative results clearly demonstrate the superiority of our method (Detail++) in mitigating semantic overflow, handling complex semantic binding, and preventing style blending, which is consistent with the quantitative results reported in Table~\ref{table:quantitative results}.

\noindent\rv{\textbf{Performance Comparison Under Different Numbers of Attributes.} We evaluate several recent approaches under prompts containing different numbers of attributes. Since most prompts in T2I-CompBench++ contain only two attributes, we first retain 600 two-attribute prompts from T2I-CompBench++\cite{huang2023t2i} (200 color, 200 texture, and 200 shape). To evaluate more challenging cases, we additionally pool prompts from ABC-6K\cite{feng2022training} and CC-500\cite{feng2022training}, and filter this pool to form two harder subsets containing 600 prompts with exactly three attributes and 600 prompts with exactly four attributes, respectively. Concretely, we use GPT-5 to parse each prompt into attribute spans according to the template in Supplementary Sec.~II, while simultaneously recording the attribute count for filtering. Using this curated test set, we evaluate Detail++ (Flux), Detail++ (SD3), Flux.1-dev, R-Bind~\cite{zhang-wan-2025-r}, T2I-R1~\cite{jiang2025t2i}, TACA (Flux)~\cite{lv2025rethinking}, and ToMe (SDXL)~\cite{hu2025token}. As shown in Fig.~\ref{fig:attribute_performance}, Detail++ effectively preserves performance as the number of attributes increases, whereas other methods are more sensitive to attribute count and exhibit noticeable degradation.}

\noindent\rv{\textbf{Inference Overhead under Different Numbers of Attributes.} 
Here, we compare inference VRAM usage and latency under three, four, and five attributes. Specifically, we measure VRAM and time consumption when generating $1024\times1024$ images. For Detail++ (SDXL), Detail++ (SD3), Flux, SD3~\cite{esser2024scaling}, R-Bind~\cite{zhang-wan-2025-r}, ToMe~\cite{hu2025token}, and SDXL~\cite{podell2023sdxl}, we set the number of inference steps to 30, while for Detail++ (Flux) we use 8 inference steps. 
All methods infer in half precision.  As shown in Fig.~\ref{fig:efficiency_comparison}, T2I-R1 exhibits the highest memory and time consumption, \rvtwo{as it is a reasoning-based method on a unified model~\cite{wu2025janus}}. In contrast, even with five branches, Detail++ (SDXL) and Detail++ (SD3) maintain substantially lower absolute VRAM usage and inference time than ToMe (SDXL) and R-Bind (SD3), respectively. We further compare the relative increases in VRAM and inference time with respect to each baseline. As reported in Table~\ref{tab: efficiency increase}, Detail++ incurs markedly smaller efficiency overhead than other training-free methods. As the flux-version Detail++ is implemented on an 8-step distilled schnell version, we can generate a high-quality image with much less inference time.
\rvtwo{It is also worth noting that Detail++ requires explicit attention maps only during the first $S$ denoising steps for self-attention map sharing and subject mask extraction.
After the branch scheduler prunes the redundant branches at step $S$, the remaining $T-S$ steps can run with fused SDPA.}}

\begin{table*}[t]
  \centering
  \caption{\rvtwo{Ablation on the self-attention propagation step $S$, reported as the fraction $S/T$ of the total denoising steps.
  Each block reports VRAM (GB), inference time (s), and mean BLIP-VQA score averaged over the color, texture, and shape subsets of T2I-CompBench++.
  For SDXL and SD3, $S/T \in \{0.0, 0.2, 0.5, 0.8, 1.0\}$; for Flux.1 Schnell, $S/T \in \{0.0, 0.25, 0.5, 0.75, 1.0\}$.
  \textbf{Bold} rows indicate the recommended $S$ for each backbone (or the nearest shown value).}}
  \label{tab:ablation-S}
  \small
  \setlength{\tabcolsep}{5pt}
  \renewcommand{\arraystretch}{1.15}
  \resizebox{0.86\textwidth}{!}{%
  \begin{tabular}{@{}c|ccc|ccc||c|ccc@{}}
  \toprule
  \multirow{2}{*}{$S/T$}
    & \multicolumn{3}{c|}{Detail++ (SDXL)}
    & \multicolumn{3}{c||}{Detail++ (SD3)}
    & \multirow{2}{*}{$S/T$}
    & \multicolumn{3}{c}{Detail++ (Flux.1 Schnell)} \\
  \cmidrule(lr){2-4}\cmidrule(lr){5-7}\cmidrule(lr){9-11}
   & VRAM & Time & BLIP-VQA & VRAM & Time & BLIP-VQA & & VRAM & Time & BLIP-VQA \\
  \midrule
  0.0  & 9.5\,GB  & 4\,s  & 0.4157 & 13.0\,GB & 14\,s & 0.4815 & 0.00  & 35.7\,GB & 5\,s  & 0.4729 \\
  0.2  & 10.9\,GB & 6\,s  & 0.5391 & \textbf{15.4\,GB} & \textbf{16\,s} & \textbf{0.7566} & 0.25  & \textbf{37.5\,GB} & \textbf{9\,s}  & \textbf{0.7620} \\
  \textbf{0.5}  & \textbf{13.1\,GB} & \textbf{10\,s} & \textbf{0.6736} & 19.0\,GB & 19\,s & 0.7822 & 0.50  & 39.4\,GB & 13\,s & 0.7634 \\
  0.8  & 15.2\,GB & 14\,s & 0.6792 & 22.6\,GB & 22\,s & 0.7856 & 0.75  & 41.2\,GB & 17\,s & 0.7649 \\
  1.0  & 16.6\,GB & 16\,s & 0.6648 & 25.0\,GB & 24\,s & 0.7728 & 1.00  & 43.1\,GB & 21\,s & 0.7533 \\
  \bottomrule
  \end{tabular}%
  }
  \end{table*}

\begin{figure}[t]
  \centering
  \includegraphics[width=0.98\linewidth]{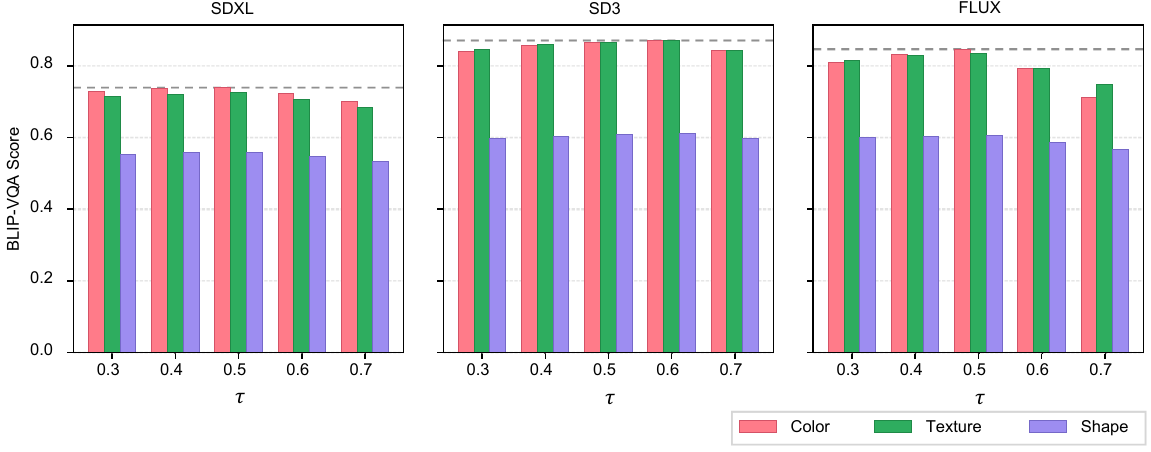}
  \caption{\rv{Ablation study of the mask binarization threshold $\tau$ across different baselines. \rvtwo{The dashed lines indicate the highest score for each backbone. All three backbones exhibit a performance plateau within $\tau \in [0.4, 0.6]$, with SD3 peaking at $\tau=0.6$ and Flux/SDXL peaking at $\tau=0.5$. The marginal differences suggest that $\tau$ is not a sensitive hyperparameter.}}}
  \label{fig:ablation-tau-selection}
\end{figure}

\begin{figure}[t]
  \centering
  \includegraphics[width=0.84\linewidth]{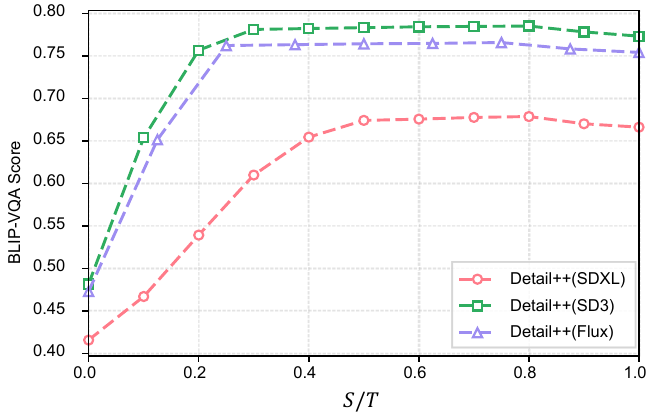}
  \caption{\rvtwo{Ablation study of the self-attention propagation step $S$ across different backbones. Each point reports the mean BLIP-VQA score averaged over color, texture, and shape subsets. DiT-based models (SD3, Flux) saturate at $S\!=\!0.25T$, while the U-Net-based model (SDXL) saturates at $S\!=\!0.5T$.}}
  \label{fig:ablation-S-selection}
\end{figure}

\subsection{Ablation Study.}
\textbf{Test-time Optimization.} As shown in Table ~\ref{tab:ablation}, we conducted a quantitative analysis of loss terms. It can be observed that using only $L_\text{ent}$~\cite{hu2025token} results in a slight improvement in the experimental outcomes. However, when the proposed $L_\text{align}$ is added, the results show a significant enhancement. Furthermore, we visualize the generated binary masks of subjects when generating the prompt ``a dog wearing sunglasses and a cat wearing a necklace'' in Fig. ~\ref{fig:ablation-ca-map}. In the middle set of images, we observe that adding the entropy loss helps the cross-attention map converge to some extent. However, the highlighted region for the `cat' token still includes the `dog's forehead and neck area, which can lead to imprecise attribute addition. This is evident in the second row of Fig. ~\ref{fig:ablation-results}, where adding the necklace to the cat also inadvertently affects the dog's neck area. Our proposed centroid alignment loss, however, further refines the initial branch generation, making the subjects' cross-attention maps more focused and accurate, greatly reducing the overlap in binary masks across different subjects, which also can be found the generated image in Fig.~\ref{fig:ablation-results}.

\rv{\textbf{Self-attention Map Propagation Steps.}
We quantitatively evaluate how the choice of the self-attention propagation step $S$ affects T2I-CompBench++ scores across different baselines.}
\rv{As shown in Fig.~\ref{fig:ablation-S-selection}, for Detail++ (SDXL), performance improves steadily as $S$ increases up to $0.5T$ and then saturates, whereas Detail++ (SD3) and Detail++ (Flux) achieve their best performance at around $S=0.25T$ and remain largely stable beyond that point.}
\rvtwo{The earlier saturation of DiT-based models can be attributed to their use of rectified flow~\cite{liuflow,esser2024scaling,flux2024}, which is explicitly designed to straighten the sampling trajectories and thereby resolves the global spatial structure in fewer initial denoising steps.
In contrast, SDXL~\cite{podell2023sdxl} requires more propagation steps to stabilize the layout across branches.
Furthermore, as shown in Table~\ref{tab:ablation-S}, increasing $S$ beyond the saturation point incurs substantial additional VRAM and inference time with negligible quality gain.
This efficiency cost stems directly from the branch scheduler, which retains all branches until step $S$ and prunes all but the last one only after the global layout has been established.
Consequently, a larger $S$ prolongs the multi-branch phase, increasing both VRAM usage and inference latency before the single-branch continuation begins.
For instance, raising $S$ from $0.5T$ to $0.8T$ on SDXL increases VRAM by 16\% and latency by 40\% while improving the mean BLIP-VQA score by only 0.005.
We therefore set $S$ at the saturation point for each backbone to achieve the best quality--efficiency trade-off, adopting $S\!=\!0.25T$ for DiT-based models and $S\!=\!0.5T$ for SDXL.}

\rv{We further conduct a qualitative analysis by visualizing intermediate branch outputs to study how $S$ influences generation under different baseline architectures.
\rvtwo{Specifically, we generate the prompt ``a blue bench and a red car'' with the branch scheduler disabled, so that the output of every branch is retained and visible.
In each group of images, reading counter-clockwise, the top-right image is branch~0 conditioned on the plain prompt ``a car and a bench'', the bottom-right is branch~1 conditioned on ``a red car and a bench'', and the large left image is the final branch conditioned on the full prompt ``a red car and a blue bench''.
As shown in Fig.~\ref{fig:ablation-n-self}, when $S=0$, the final branch produces a layout that deviates substantially from that of branch~0, leading to attribute misassignment.
As $S$ increases, the layout inconsistencies highlighted by the red bounding boxes progressively diminish, indicating that self-attention map sharing enforces a more coherent spatial structure across branches.
At the extreme case of $S\!=\!T$, however, the last branch may exhibit degraded texture in attribute-modified regions.
For instance, the blue coloring on the bench surface appears unnatural, as the shared layout signal persists throughout the entire denoising process and leaves insufficient steps for local detail refinement.
This observation is consistent with the slight downward trend in BLIP-VQA scores at large $S$ reported in Fig.~\ref{fig:ablation-S-selection}.}}

\rv{\textbf{Inference Optimization Components.} In this section, we evaluate the impact of each inference optimization component when using Detail++ under three branches. As shown in Table~\ref{tab:eff_ablation}, enabling \emph{batch expansion} leads to a consistent reduction in both average VRAM usage and inference latency across backbones: for Detail++ (SDXL), VRAM drops by 13.54\% and latency by 5.88\%; for Detail++ (FLUX), VRAM drops by 8.00\% and latency by 8.70\%. Enabling the \emph{branch scheduler} further amplifies these gains, yielding substantial reductions in both VRAM and latency (SDXL: 32.03\% VRAM and 41.18\% latency; FLUX: 19.94\% VRAM and 60.87\% latency).}
\rvtwo{We also observe that both batch expansion and the branch scheduler incur only minor quality degradation.
For batch expansion, the measured shift is primarily attributable to subtle numerical differences between batched and non-batched tensor computations, and is negligible in practice.
For the branch scheduler, which retains only the last branch and completes the remaining denoising steps with the full prompt in isolation, the performance loss remains limited.
We attribute this to accumulative latent modification, which has already explicitly injected each attribute into the latent regions associated with the corresponding subjects before redundant branches are pruned.
Because the subject-attribute structure is already encoded in the latent, the subsequent low-noise refinement steps mainly adjust local appearance and do not substantially alter the global semantic layout.}

\rv{\textbf{Mask Binarization Threshold.} We quantitatively evaluate how the choice of cross-attention map binarization threshold $\tau$ affects T2I-CompBench++ scores across different backbones. As shown in Fig.~\ref{fig:ablation-tau-selection}, 
\rvtwo{all three backbones exhibit a broad performance plateau within $\tau \in [0.4, 0.6]$.
Detail++ (SD3) attains its best score at $\tau=0.6$, while Detail++ (Flux) and Detail++ (SDXL) peak at $\tau=0.5$, yet the score differences across this range remain marginal, below 0.04 in averaged BLIP-VQA.
This robustness indicates that $\tau$ is not a sensitive hyperparameter requiring per-model tuning, and we adopt $\tau=0.5$ as a robust default across all backbones.}}

\subsection{User Study}
\begin{figure}[t]
    \centering
    \includegraphics[width=0.70\linewidth]{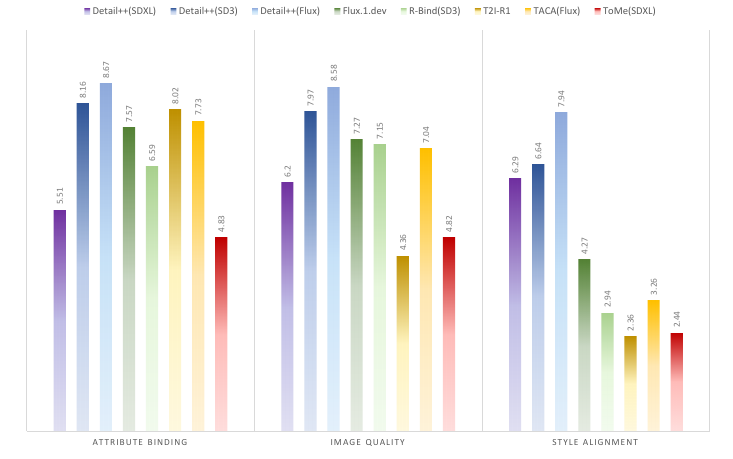}
    \caption{User study results evaluating different methods across three key metrics. Detail++ consistently outperforms all baseline methods in attribute binding, image quality, and style alignment, receiving significantly higher scores particularly in style alignment.}
    \vskip -0.4cm
    \label{fig:user_study}
\end{figure}

To comprehensively evaluate the effectiveness of our proposed method, we conducted a user study with 145 participants (Fig.~\ref{fig:user_study}). Each questionnaire contained 12 questions spanning three key aspects: \emph{Attribute Binding}, \emph{Image Quality}, and \emph{Style Binding}. \rv{To ensure reliable and unbiased evaluation, we constructed an image pool for each method---ToMe~\cite{hu2025token}, R-Bind~\cite{zhang-wan-2025-r}, T2I-R1~\cite{jiang2025t2i}, TACA~\cite{shin2025exploring}, and our Detail++}---where each pool contained 120 images generated from 30 diverse prompts using four different random seeds. For each survey question, images were randomly sampled from these pools according to the corresponding prompt. The results indicate that Detail++ achieves strong attribute binding and style composition while maintaining competitive image quality.

\vspace{-4pt}
\section{Conclusion}
We present Detail++, a training-free approach for accurate detail binding in text-to-image generation with complex prompts. Our Progressive Detail Injection framework combines self-attention map sharing and Accumulative Latent Modification to ensure proper attribute assignment while maintaining consistent layouts. The Centroid Alignment Loss further improves binding precision via cross-attention map optimization. Extensive experiments demonstrate that Detail++ significantly outperforms existing methods in preventing attribute overflow, mismatching, and style blending while preserving high image fidelity. As a plug-and-play module compatible with current diffusion models, Detail++ represents an important step toward more controlled and semantically accurate text-to-image generation without additional training.

\vspace{-4pt}
\section*{Acknowledgement}
This work was supported by the National Natural Science Foundation of China (No. 6250070674) and the Zhejiang Leading Innovative and Entrepreneur Team Introduction Program (2024R01007).

%% file: sec/supplementary.tex
\clearpage
\setcounter{page}{1}
\maketitlesupplementary

\setcounter{equation}{0}
\setcounter{table}{0}
\setcounter{figure}{0}
\renewcommand\thefigure{A\arabic{figure}}
\renewcommand\thetable{A\arabic{table}}
\renewcommand\theequation{A\arabic{equation}}

\section{Details and Analysis}
\label{app:details and analysis}
\subsection{Limitation and Discussion}
\label{app:limitation and discussion}
The proposed Progressive Detail Injection (PDI) framework addresses detail binding issues in text-to-image generation in a training-free manner, yet it has several limitations. First, the method relies heavily on the quality and accuracy of the initial self-attention maps. If the early-stage layout is suboptimal, subsequent attribute injection may not fully correct the errors, which can ultimately limit the final generation quality. \rv{This issue can be substantially alleviated by more advanced text-to-image model architectures. For instance, for the prompt ``a bench and a car'', SDXL often struggles to produce a correct composition and may exhibit subject blending, merging the two entities into one. In contrast, DiT-based models can typically separate the two subjects clearly. Moreover, by visualizing cross-attention maps during generation, we observe that Flux produces more disentangled and concentrated attention for the two subjects, whereas SDXL exhibits substantial overlap and more diffuse activations.}

\begin{figure}[h]
  \centering
  \includegraphics[width=0.9\linewidth]{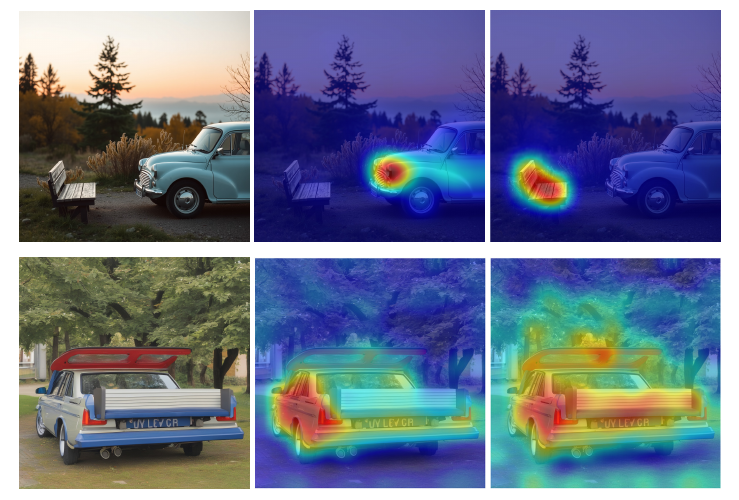}
  \caption{\rv{Comparison of multi-subject generation. The first row shows Flux results, and the second row shows SDXL results.}}
  \label{fig:dit-unet-comparison}
\end{figure}

\subsection{Experiment Details}
\label{app:experiment details}

\noindent\textbf{Detail++ Implementation Details.} We implement our method on \rv{SDXL, SD3-Medium, and Flux.1-schnell}. For SDXL, we extract cross-attention maps from all layers at a resolution of $32\times32$ across all U-Net blocks, as we find this setting provides comparatively fine-grained and accurate localization. \rv{For DiT-based models, following the findings in~\cite{shin2025exploring}, we dynamically select the top-5 layers with the highest activation values.} All experiments are conducted on a single NVIDIA H20 GPU.


\noindent\textbf{Baseline Implementation Details.} For quantitative comparison, we use official implementation of Stable diffusion 1.5~\cite{rombach2022high}, Stable diffusion XL~\cite{podell2023sdxl}, Structured Diffusion Guidance~\cite{feng2022training}, Composable diffusion~\cite{liu2022compositional}, PixelArt-$\alpha$~\cite{chen2023pixart}, ELLA~\cite{hu2024ella}, ToME~\cite{hu2025token}, Attention Regulation~\cite{zhang2024enhancing},\rv{ IterComp\cite{zhangitercomp},R-Bind\cite{zhang-wan-2025-r},T2I-R1\cite{jiang2025t2i} and TACA\cite{lv2025rethinking}. Since the SDXL checkpoint of Ranni~\cite{feng2024ranni} is not open-sourced, thus we directly refer to the score in their paper. For TACA\cite{lv2025rethinking}, we use its lora-finetuned version; For RPG\cite{yang2024mastering}, we use the state-of-art llm planner, i.e. GPT-5, with an revised llm prompt.}

\rvtwo{\noindent\textbf{Quantitative Results Analysis.}
We further analyze the experimental results in Table~1.
It can be observed that Detail++ scores notably lower on the shape dimension compared to color and texture.
We attribute this to two reasons.
First, across all three backbones (SDXL, SD3, and Flux.1), the baseline shape scores are inherently lower than those for color and texture, which shows that accurate shape generation is an inherently difficult problem for text-to-image generation.
Second, Detail++ has relatively limited capacity to modify subject shape, as shape is a structural property of the image that is largely encoded in the self-attention maps, which are shared across all subsequent branches.}

\begin{figure}[h]
  \centering
  \includegraphics[width=0.92\linewidth]{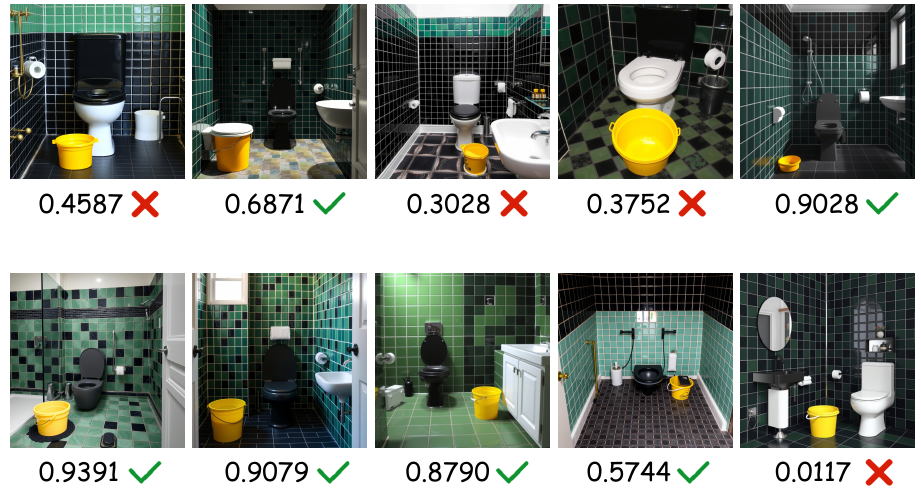}
  \caption{\rvtwo{Qualitative stability comparison for the prompt ``A black and green tile bathroom with a black toilet and a yellow bucket on the floor.'' The first and second rows show images generated by Flux.1-dev and Detail++ (Flux.1), respectively, under identical inference settings. The score beneath each image is the product of the BLIP-VQA attribute binding scores for ``a black and green tile bathroom'', ``a black toilet'', and ``a yellow bucket''.}}
  \label{fig:stability}
\end{figure}

\rvtwo{\noindent\textbf{Generation Stability Analysis.}
To quantify the stability of Detail++ (Flux.1), we conduct both quantitative and qualitative analyses on Flux.1 Schnell.
Specifically, we randomly select 150 prompts from T2I-CompBench++~\cite{huang2023t2i} and generate 5 images per prompt under different random seeds, yielding 750 generation attempts per method.
A generation is considered ``successful'' if the product of all per-attribute BLIP-VQA scores exceeds 0.5.
As shown in Table~\ref{tab:stability}, Detail++ (Flux.1) achieves a per-attempt success rate of 71\%, compared with 48\% for the Flux.1-dev baseline.
Fig.~\ref{fig:stability} further provides a qualitative illustration.
Across 5 random seeds, Detail++ (Flux.1) is more likely to produce images that fully align with the prompt, whereas the Flux.1-dev baseline frequently exhibits attribute mismatches or attribute blending.}

\begin{table}[h]
  \caption{\rvtwo{Generation stability of Detail++ (Flux.1) and the Flux.1-dev baseline on 150 T2I-CompBench++ prompts (5 seeds each). SR is the fraction of attempts in which the product of all per-attribute BLIP-VQA scores exceeds 0.5.}}
  \label{tab:stability}
  \centering
  \small
  \setlength{\tabcolsep}{5pt}
  \renewcommand{\arraystretch}{1.1}
  \resizebox{\columnwidth}{!}{%
  \begin{tabular}{lccc}
    \toprule
    \textbf{Method} & \textbf{SR (\%) $\uparrow$} & \textbf{Failure Rate (\%) $\downarrow$} & \textbf{Avg. attempts $\downarrow$} \\
    \midrule
    Flux.1-dev        & 48.0 & 52.0 & 2.1 \\
    Detail++ (Flux.1) & \textbf{71.0} & \textbf{29.0} & \textbf{1.4} \\
    \bottomrule
  \end{tabular}%
  }
\end{table}

\rvtwo{\noindent\textbf{Ablation Study Setup.}
Generally, the ablation studies in Sec.~V-C adopt the evaluation metrics and test splits defined in T2I-CompBench++~\cite{huang2023t2i}.
We evaluate on its three standard attribute-binding subsets separately, namely color, texture, and shape.
For each subset, we use 300 prompts taken from the test portion of T2I-CompBench++.
For each prompt, we generate five independent samples with different random seeds and report the average score.
All generations use a resolution of $1024\!\times\!1024$.
We set the number of denoising steps to 30 for Detail++ (SDXL) and Detail++ (SD3), and to 8 for Detail++ (Flux), which is built on Flux.1 Schnell.}

\subsection{Sub-prompts Formulations}
\label{app:sps management}
In this section, we compare the effects of different prompt formulations granularities on generation quality. For example, given the original prompt \( p_{\text{ori}} =  \)\textit{“a red dog with sunglasses and a blue cat with a necklace”}, we can manage the sub-prompts in two ways: 1) most complex first, or 2) simplest first. Regarding granularity, we can manage them in two ways: a) one subject per branch, or b) one attribute per branch. This results in a total of four possible configuration combinations. These are:

\textbf{Config. A (1 + a):}
\[
\mathcal{P} = \left\{
\begin{array}{l}
    p_0 : \textit{“a red dog with sunglasses and a blue cat}\\ \textit{with a necklace.”}, \\
    p_1 : \textit{“a dog and a cat.”}, \\
    p_2 : \textit{“a red dog with sunglasses and a cat.”}, \\
    p_3 : \textit{“a dog and a blue cat with a necklace.”} 
\end{array} \right\},
\]
\[
\mathcal{Q} = \left\{
\begin{array}{l}
    q_1 : \textit{“dog”}, \\
    q_2 : \textit{“cat”} 
\end{array} \right\}.
\]

\textbf{Config. B (1+b):}
\[
\mathcal{P} = \left\{
\begin{array}{l}
    p_0 : \textit{“a red dog with sunglasses and a blue cat}\\ \textit{with a necklace.”}, \\
    p_1 : \textit{“a dog and a cat.”}, \\
    p_2 : \textit{“a red dog and a cat.”}, \\
    p_3 : \textit{“a dog with sunglasses and a cat.”}, \\
    p_4 : \textit{“a dog and a blue cat.”},\\
    p_5 : \textit{“a dog and a cat with a necklace.”}
\end{array} \right\},
\]
\[
\mathcal{Q} = \left\{
\begin{array}{l}
    q_1 : \textit{“dog”}, \\
    q_2 : \textit{“dog”}, \\
    q_3 : \textit{“cat”}, \\
    q_4 : \textit{“cat”}
\end{array} \right\}.
\]

\textbf{Config. C (2 + a):}
\[
\mathcal{P} = \left\{
\begin{array}{l}
    p_1 : \textit{“a dog and a cat.”}, \\
    p_2 : \textit{“a red dog with sunglasses and a cat.”}, \\
    p_3 : \textit{“a dog and a blue cat with a necklace.”} 
\end{array} \right\},
\]
\[
\mathcal{Q} = \left\{
\begin{array}{l}
    q_1 : \textit{“dog”}, \\
    q_2 : \textit{“cat”} 
\end{array} \right\}.
\]

\textbf{Config. D (2+b)}:
\[
\mathcal{P} = \left\{
\begin{array}{l}
    p_1 : \textit{“a dog and a cat.”}, \\
    p_2 : \textit{“a red dog and a cat.”}, \\
    p_3 : \textit{“a dog with sunglasses and a cat.”}, \\
    p_4 : \textit{“a dog and a blue cat.”},\\
    p_5 : \textit{“a dog and a cat with a necklace.”}
\end{array} \right\},
\]
\[
\mathcal{Q} = \left\{
\begin{array}{l}
    q_1 : \textit{“dog”}, \\
    q_2 : \textit{“dog”}, \\
    q_3 : \textit{“cat”}, \\
    q_4 : \textit{“cat”}
\end{array} \right\}.
\]

\begin{table*}[t]
  \caption{Inference consumption and quality of different sub-prompt formulation configurations \rv{for Detail++ with different backbones}. We test BLIP-VQA scores of color, texture and shape in T2I-CompBench++ with efficient self-attention propagation enabled.}
  \centering
  \setlength{\tabcolsep}{11pt}
  \begin{tabular}{c c c c | c c c | c c}
    \toprule
    \multirow{2}{*}{Backbone} & \multirow{2}{*}{Config} & \multirow{2}{*}{1/2} & \multirow{2}{*}{a/b} &
    \multicolumn{3}{c|}{BLIP-VQA} &
    \multirow{2}{*}{\makecell{Time\\Consumption}} &
    \multirow{2}{*}{\makecell{Memory\\Consumption}} \\
    & & & & Color & Texture & Shape & & \\
    \midrule

    \multirow{5}{*}{Detail++(SDXL)}
      & $A$ & 2 & b & 0.6750 & 0.6620 & 0.5150 & 13s& 14.75G\\
      & $B$ & 1 & b & \bluenum{\textbf{0.7389}} & \bluenum{\textbf{0.7241}} & \bluenum{\textbf{0.5582}} & 17s& 18.31G\\
      & $C$ & 2 & a & 0.6300 & 0.6050 & 0.4700 & 10s& 13.05G\\
      & $D$ & 1 & a & 0.7250 & 0.7120 & 0.5480 & 1s& 15.78G\\
      & \multicolumn{3}{c|}{accumulative prompt} & 0.7000 & 0.6880 & 0.5300 & 15s& 15.78G\\
    \midrule

    \multirow{5}{*}{Detail++(SD3)}
      & $A$ & 2 & b & \bluenum{\textbf{0.8703}} & \bluenum{\textbf{0.8699}} & \bluenum{\textbf{0.6100}} & 24s& 26.31G\\
      & $B$ & 1 & b & 0.7800 & 0.7780 & 0.5450 & 32s& 27.79G\\
      & $C$ & 2 & a & 0.8450 & 0.8420 & 0.5900 & 21s& 25.88G\\
      & $D$ & 1 & a & 0.7300 & 0.7280 & 0.5050 & 28s& 27.23G\\
      & \multicolumn{3}{c|}{accumulative prompt} & 0.8200 & 0.8180 & 0.5700 & 28s& 27.23G\\
    \midrule

    \multirow{5}{*}{Detail++(Flux)}
      & $A$ & 1 & a & \bluenum{\textbf{0.8398}} & \bluenum{\textbf{0.8305}} & \bluenum{\textbf{0.6032}} & 10.1s& 39.3G\\
      & $B$ & 1 & b & 0.7500 & 0.7400 & 0.5400 & 11.4s& 40.1G\\
      & $C$ & 2 & a & 0.8150 & 0.8060 & 0.5900 & 8.8s& 38.54G\\
      & $D$ & 2 & b & 0.7000 & 0.6900 & 0.5050 & 10.8s& 39.6G\\
      & \multicolumn{3}{c|}{accumulative prompt} & 0.7900 & 0.7800 & 0.5700 & 10.8s& 39.6G\\
    \bottomrule
  \end{tabular}
  \label{tab:subprompts_config_extended}
\end{table*}

Therefore, we conduct plenty of experiments to test which one can best reflect our method's efficiency, as shown in Table. ~\ref{tab:subprompts_config_extended}. Another sub-prompts approach is accumulative prompts, where the previously added attribute continues to appear in the prompt of the next branch. This is referred to as the
\textbf{Accumulative Prompt}:
\[
\mathcal{P} = \left\{
\begin{array}{l}
    p_1 : \textit{“a dog and a cat.”}, \\
    p_2 : \textit{“a red dog and a cat.”}, \\
    p_3 : \textit{“a red dog with sunglasses and a cat.”}, \\
    p_4 : \textit{“a red dog with sunglasses and a blue cat.”},\\
    p_5 : \textit{“a red dog with sunglasses and a blue cat} \\ \textit{a with necklace.”}
\end{array} \right\},
\]
\[
\mathcal{Q} = \left\{
\begin{array}{l}
    q_1 : \textit{“dog”}, \\
    q_2 : \textit{“dog”}, \\
    q_3 : \textit{“cat”}, \\
    q_4 : \textit{“cat”}
\end{array} \right\}.
\]
However, as shown in the second-to-last row of Table ~\ref{tab:subprompts_config_extended}, the quantification results are not ideal. We hypothesize that this is due to modifier overflow in the longer prompts ~\cite{hu2025token}, which degrades the efficiency of our method.

\begin{figure*}[h]
    \centering
    \includegraphics[width=0.9\linewidth]{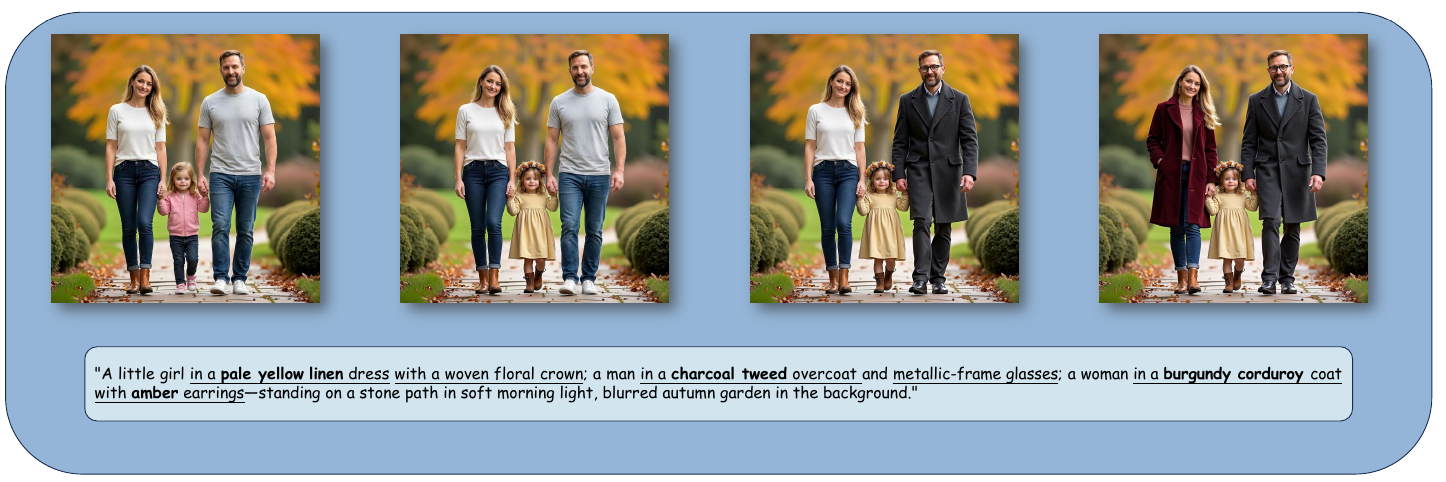}
    \caption{\rv{Visualization of generating a image with 9 attributes and 3 subjects by Detail++(Flux).}}
    \label{app: dit-subject-binding-case}
\end{figure*}

\rv{According to Table~\ref{tab:subprompts_config_extended}, we compare how different sub-prompt formulations affect generation quality, and find that for more advanced models (i.e., SD3 and Flux), the performance gap across sub-prompts becomes smaller. Notably, the degradation introduced by a more aggressive attribute-addition strategy (config.C) is no longer pronounced. As illustrated in Fig.~\ref{app: dit-subject-binding-case}, we disable the branch scheduler and visualize the generation process for a prompt containing three subjects with nine attributes in total. Under the config.C-style sub-prompt formulation, Detail++ (Flux) progressively augments ``a little girl'' with ``in a pale yellow linen dress with a woven floral crown'', augments ``man'' with ``in a charcoal tweed overcoat and metallic-frame glasses'', and augments ``woman'' with ``in a burgundy corduroy coat with amber earrings''.}

\begin{table}[t]
\centering
\caption{Elements candidates used in style composition evaluation.}
\label{tab:attr_candidates}
\small
\setlength{\tabcolsep}{6pt}
\renewcommand{\arraystretch}{1.15}
\begin{tabularx}{0.95\linewidth}{l X}
\toprule
\textbf{Attribute} & \textbf{Candidates} \\
\midrule
Type of style (14) &
Anime, ukiyo-e, ink painting, Lego, oil painting, cyberpunk, sketch, pixel art,
watercolor, graffiti, low-poly, isometric, charcoal drawing, chalk. \\

Subject (16) &
\textit{Animal:} dog, cat, horse, panda, tiger, unicorn;
\textit{Character:} robot, astronaut, woman, man;
\textit{Object:} spaceship, building, bicycle, train, piano, car. \\

Background (10) &
Forest, space, desert, city, ruins, ocean, marketplace, volcano, swamp, factory. \\
\bottomrule
\end{tabularx}
\end{table}

\begin{figure*}[t]
    \centering
    \includegraphics[width=0.85\linewidth]{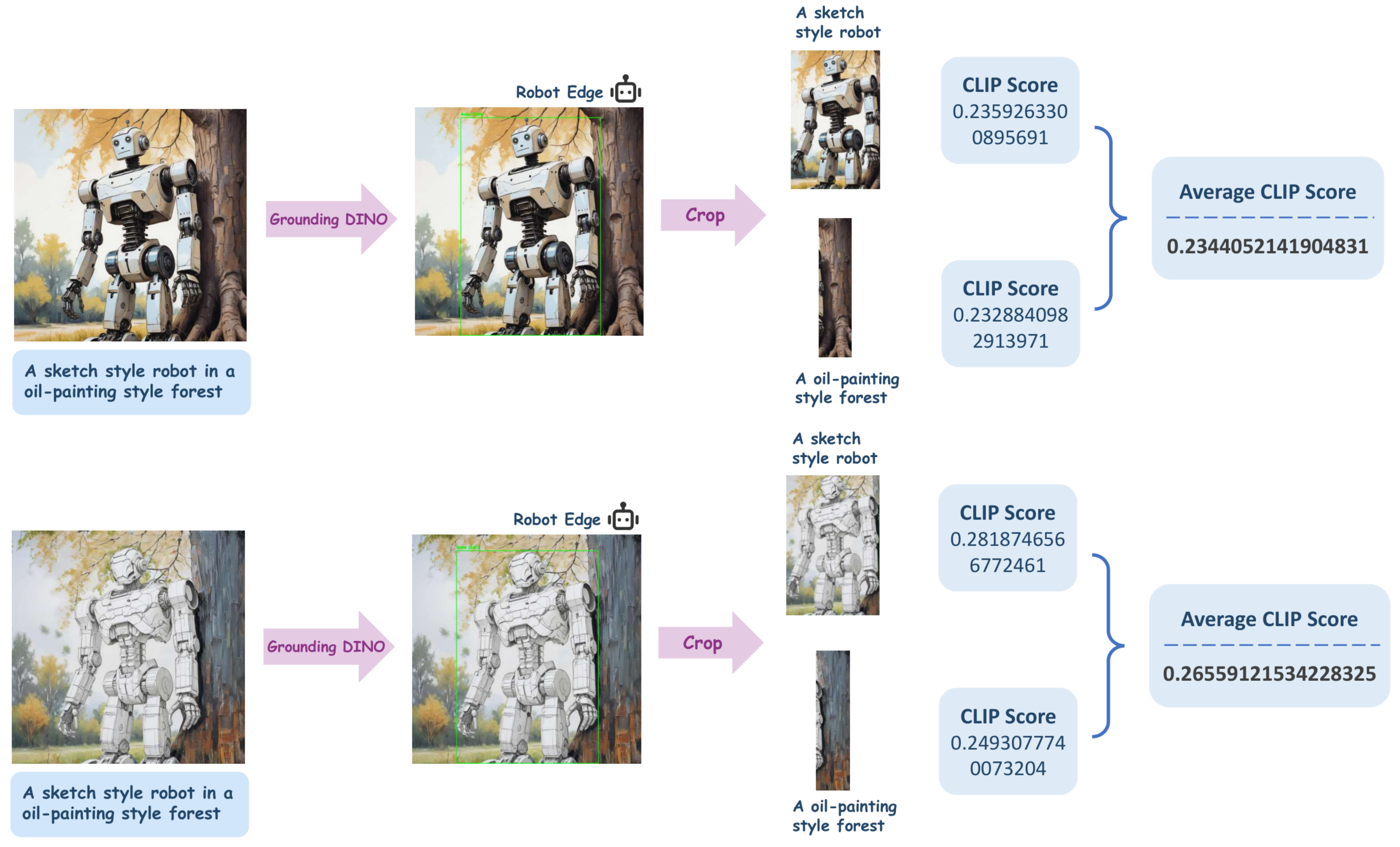}
    \caption{Pipeline of measure the style composition accuracy.}
    \label{app:fig_scb_pipeline}
\end{figure*}
\subsection{SCB details}
\label{app:benchmark construction}
\noindent\textbf{Style Composition Benchmark (SCB) construction.} To comprehensively evaluate text-to-image generation under multi-style composition scenarios,\rv{we construct SCB prompts using three templates:
``A [style1] style [subject1] in a [style2] style [background].'',
``A [style1] style [subject1] next to a [style2] style [subject2].'',
and ``A [style1] [subject1] and a [style2] [subject2] in a [style3] [background].''} 
As shown in Table~\ref{tab:attr_candidates}, \rv{we provide 14 candidate styles, 16 candidate subjects, and 10 candidate backgrounds.} We enforce that styles assigned within a single prompt are pairwise distinct, and that multi-subject prompts do not reuse the same subject. \rv{We then deduplicate the resulting text prompts to ensure that the final benchmark contains exactly 1{,}000 unique prompts.}

\noindent\textbf{SCB evaluation metrics.}\label{App:SCB evaluation} 
We propose a novel style alignment evaluation methodology that employs object detection to segment composite images into subject and background components. Each segment is independently analyzed against its intended style description using CLIP score metrics. The final averaged score reflects the overall fidelity of the composite image to its intended style descriptors across all elements. As illustrated in Fig.~\ref{app:fig_scb_pipeline}, our method effectively evaluates style alignment across composite images. For upper half of the fugure, the robot incorporates oil painting stylistic elements while the forest background exhibits traces of sketch characteristics, resulting in a lower CLIP score (0.2344) that indicates reduced alignment due to style contamination. In contrast, the lower half of the figure demonstrates clearer stylistic delineation—a sketch-style robot (0.2818) against an oil painting forest background (0.2493)—yielding a higher average alignment score (0.2656). This granular approach enables precise measurement of style fidelity across different image elements. Noticed that, why we use a cropping way instead of a segmentation way, is the Clip cannot accurately recognize the images with partial vacancy.

\begin{figure*}[h]
    \centering
    \includegraphics[width=0.92\linewidth]{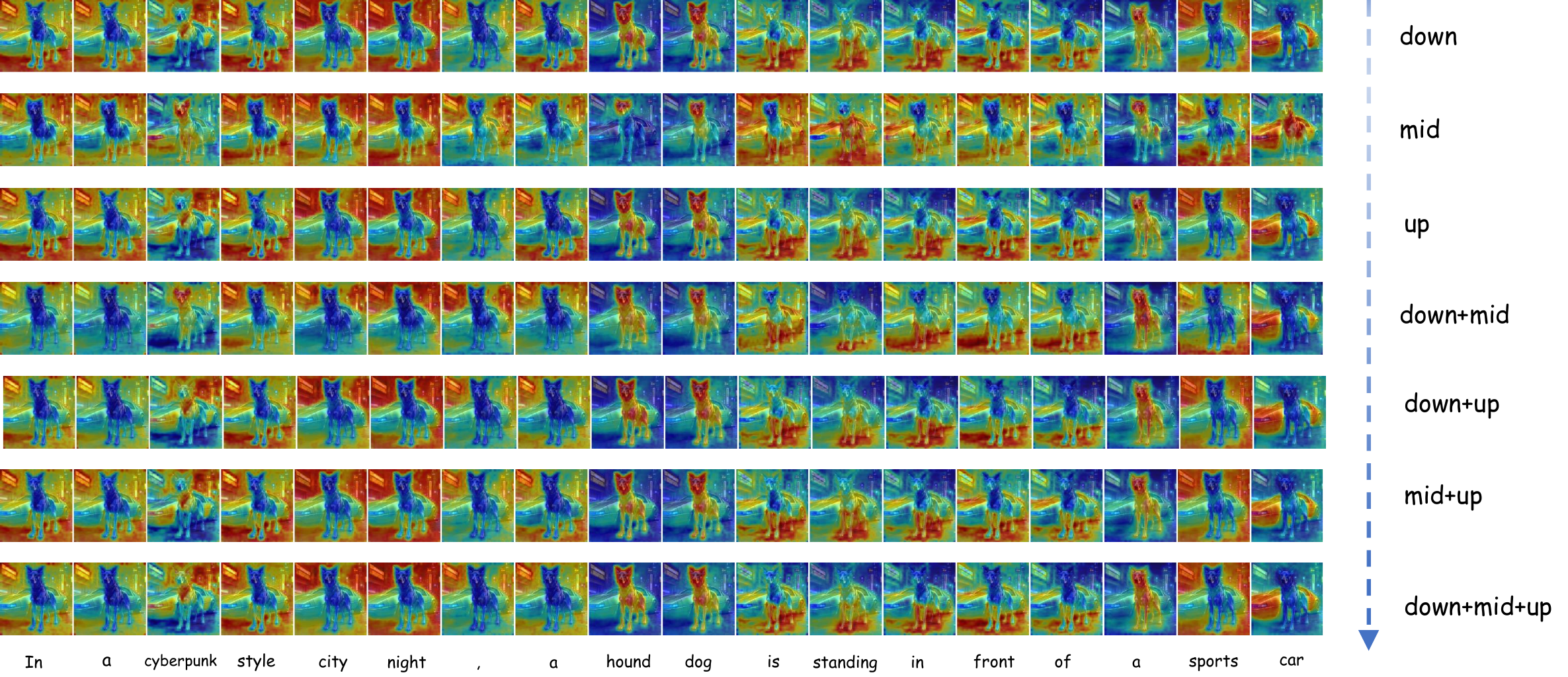}
    \caption{Visualization of cross-attention maps at different layers. “Down” denotes the cross-attention layers in the first downsampling block, while “Up” indicates those in the second upsampling block. Other layers with a 64$\times$64 cross-attention map require significantly more memory and are thus omitted here to avoid excessive GPU usage.}
    \label{app: fig-different attention selection visualization}
\end{figure*}

\begin{figure*}[t]
    \centering
    \includegraphics[width=0.9\linewidth]{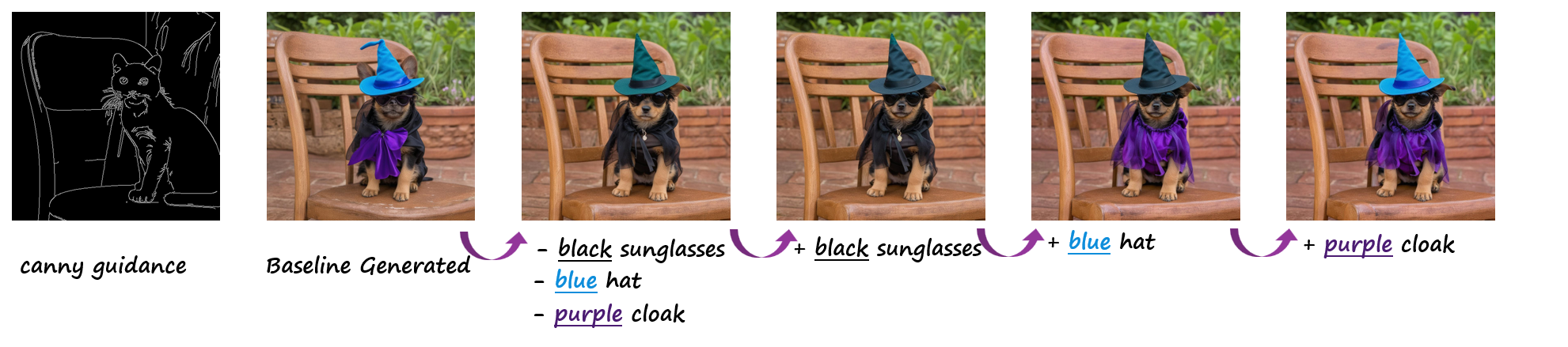}
    \caption{Our methods can generate images with accurate detail binding.}
    \label{app: with-control-net}
\end{figure*}

\begin{figure}[h]
    \centering
    \includegraphics[width=0.9\linewidth]{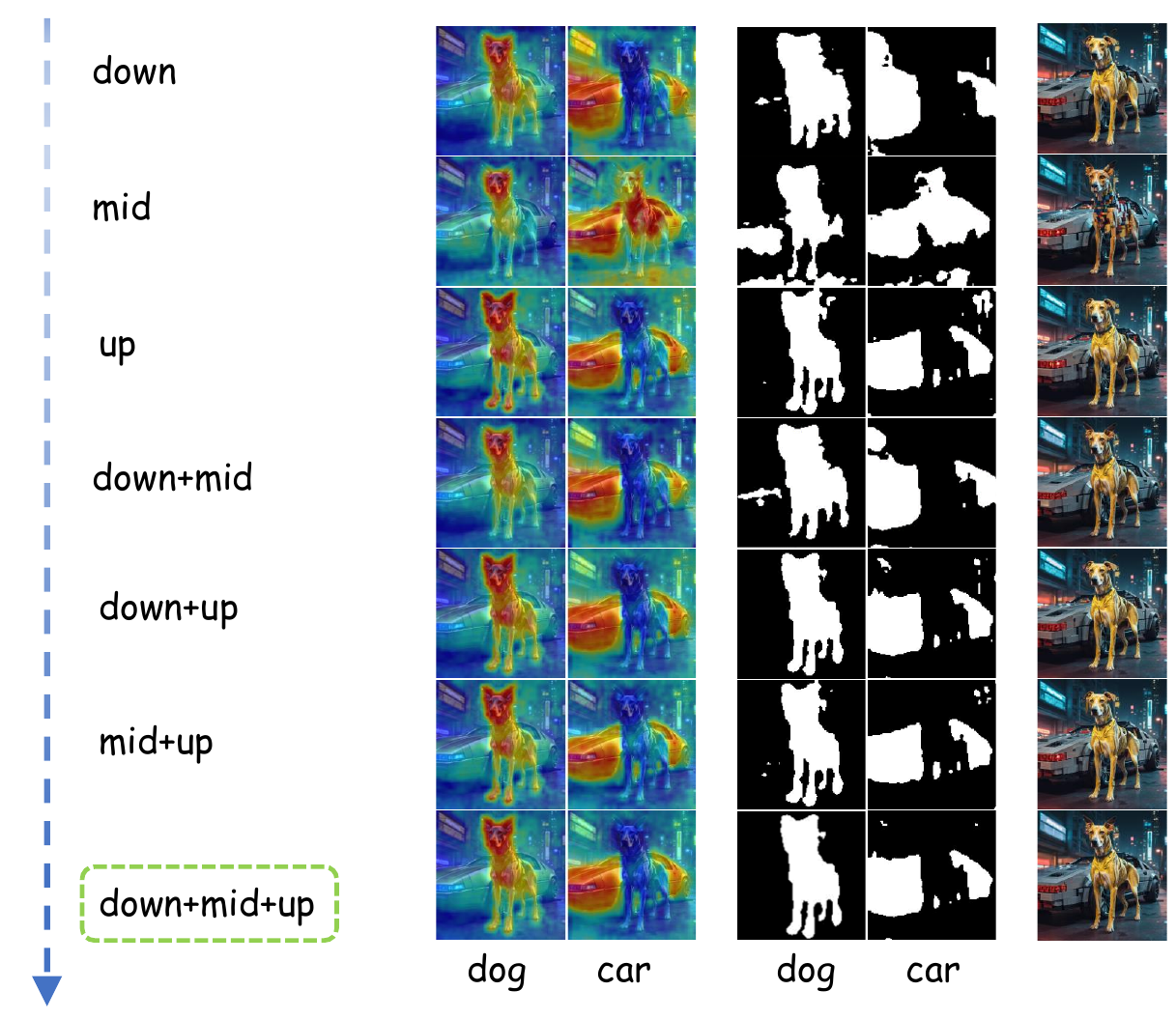}
    \caption{Comparison of the effects of different cross-attention layer selections in binary mask extraction. It can be observed that extracting the 32$\times$32 resolution cross-attention map from the down, up, and mid blocks yields the cleanest and most accurate subject mask.}
    \label{app: binary-mask selection}
\end{figure}

\begin{figure}[t]
\centering
\includegraphics[width=0.75\linewidth]{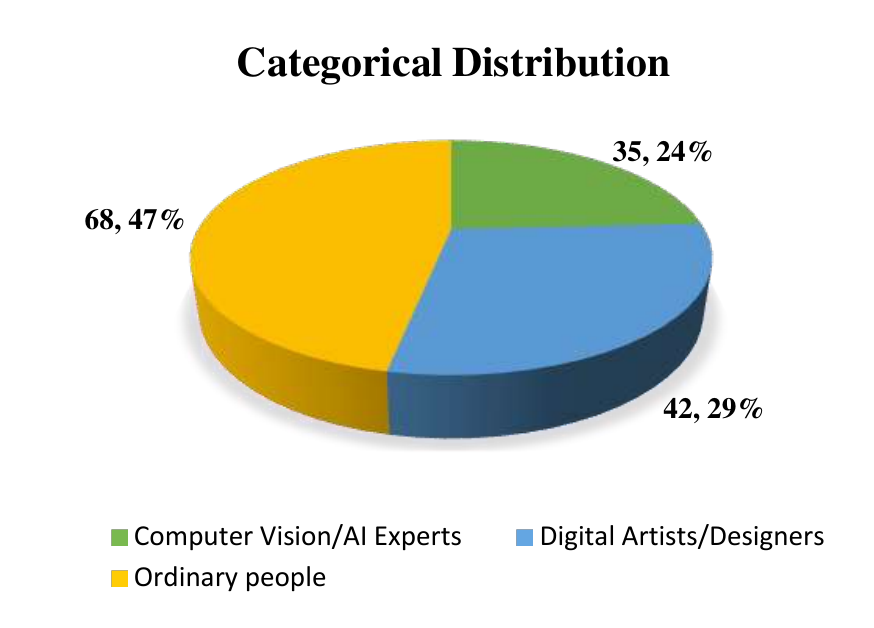}
\caption{
\rv{\textbf{Participant distribution. }The diverse participant composition, encompassing AI experts , digital artists , and general public, enables comprehensive evaluation across different dimensions, strengthening the validity of our method.}
}
\label{fig:user_2}
\end{figure}

\subsection{Comparison between SA and CA mechanism}
In this section, we present key concepts and techniques that serve as the foundation for our proposed method. These preliminary components provide a clearer understanding of the motivations behind our approach and the technical mechanisms leveraged in its design.

\begin{figure*}[t]
  \centering
  \includegraphics[width=0.9\linewidth]{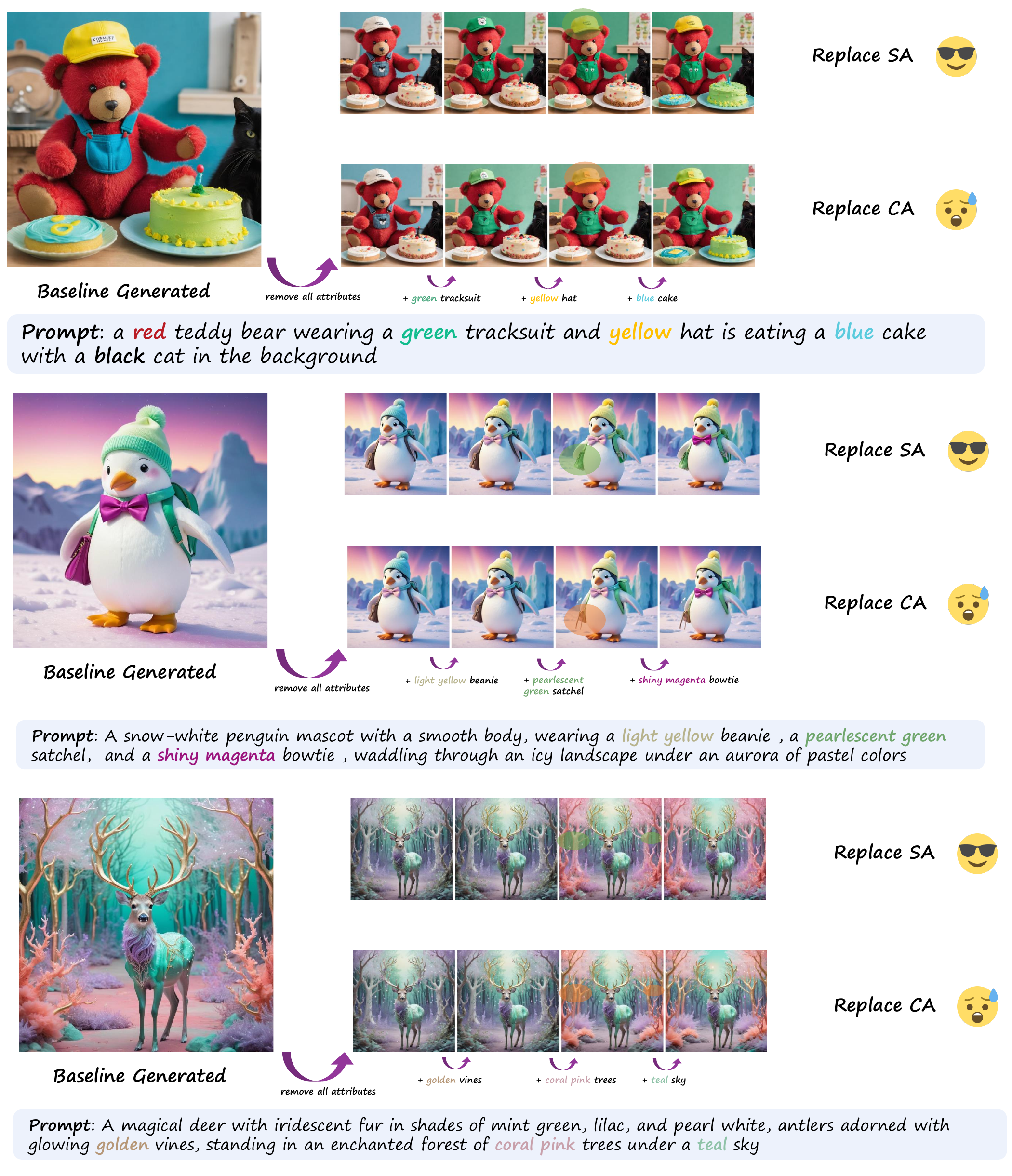}
  \caption{Comparative analysis of attention map types in image generation. As indicated by the marked areas, replacing the cross-attention map results in weaker editing effects, making it less effective at accurately assigning attributes to the subject. In contrast, replacing the self-attention map allows for more precise attribute assignment in the same scenario.}
  \label{app: sa?ca?}
\end{figure*}

\noindent\textbf{Why SA map?} While many recent studies ~\cite{nam2024dreammatcher,liu2024towards,cao2023masactrl,tumanyan2023plug} indicate that self-attention maps carry structural information and thus have strong layout-preserving capabilities, the cross-attention map-based editing approach ~\cite{hertz2022prompt} still has a significant impact. So, why choose to replace the self-attention map rather than the cross-self attention map for progressive editing? Here, we visualize the difference between replacing self-attention maps (SA) and cross-attention maps (CA) at various steps. As shown in Fig. ~\ref{app: sa?ca?}, editing based on the self-attention map tends to provide stronger editing capabilities, consistent with the findings in FPE~\cite{liu2024towards}.

We hypothesize that the relatively weaker effect of editing based on cross-attention arises because it only replaces or influences the attention map of the edited object. In contrast, editing based on the self-attention map is more like conditional generation under the prior assumption of a diffusion model’s layout, which resembles the effect described in ~\cite{zhang2023adding}. In this way, our framework, \textbf{Detail++}, ensures that each editing step is guided by an accurate detail binding relation prompt, enabling precise detail binding and preventing issues such as overflow, mismatching, and blending.

\subsection{User Study}
\noindent\rv{\textbf{Questionnaire design.} 
To build a diverse and balanced set of stimuli for the questionnaire, we curated a prompt pool spanning three evaluation dimensions: Attribute Binding, Style Binding, and Image Quality. As summarized in Table~\ref{tab:prompt_pool}, each dimension contains 12 distinct prompts (36 prompts in total). For each prompt, we generate images using 8 methods under comparison, and apply 4 different random seeds. This yields an image pool of 8×4=32 candidate images per prompt. During questionnaire construction, images are sampled from the corresponding prompt-specific pool, while keeping model identities hidden from participants. 
Each questionnaire contains 12 questions covering three dimensions. For each participant, we randomly sample 4 questions per dimension (12 questions in total) from the prompt pool in Table~\ref{tab:prompt_pool}. Each question is formulated as a multiple-choice selection task: participants are shown a set of images generated under the same prompt (from the prompt-specific image pool constructed by 8 methods × 4 seeds), and are instructed to select all images that satisfy the requirement stated in the question.}

\rv{Before the formal questionnaire, We conducted a pilot study with 15 participants. Based on their feedback, we refined the question wording, adjusted the time allocations, and clarified the evaluation criteria before proceeding with the full-scale study.}

\rv{Throughout our user study, we implemented comprehensive ethical protocols to ensure participant rights and data security. All participants underwent a structured informed consent process, clearly outlining the study objectives and their right to withdraw. Data collection and storage adhered to strict privacy protection measures, with immediate anonymization through unique identifiers and encrypted storage systems.}

\noindent\rv{\textbf{Participant demographics.} 
145 participants with diverse backgrounds and expertise levels were invited to answer the questionnaire. As showed in Fig.~\ref{fig:user_2}, the participant pool comprised three main groups: computer vision and AI experts (24.1\%), digital artists and designers (29.0\%), and general users (46.9\%). This diverse demographic composition helped minimize potential biases and ensure the generalizability of our evaluation results across different user groups and experience levels.}

\noindent\rv{\textbf{Analysis of collected results.} 
Firstly, we preprocessed the collected questionnaires, screened out unqualified questionnaires according to the average response time in the questionnaires, and finally conducted statistical analysis on three categories of questions in the qualified questionnaires. The number of votes received by each model in a particular category is \( C_{\text{model}} \). We standardized the number of votes to a ten-point system:}

\[
\text{Score}_{\text{model}} = \left( \frac{C_{\text{model}}}{\text{Total number of votes}} \right) \times 10.
\]

\rv{Through calculation, we can get the score of each model in the three aspects.}

\subsection{Cross-attention Layers Selection}
\label{app: ca layer selection}
In this section, we compare different layer selections for binary mask generation. We visualize cross-attention maps from various layers (see Fig.~\ref{app: fig-different attention selection visualization}) and provide additional detailed visualizations in Fig.~\ref{app: binary-mask selection}. It is evident that extracting the 32$\times$32 resolution cross-attention map from the down, up, and mid blocks produces the cleanest and most accurate subject mask.

\begin{figure}[t]
  \centering
  \includegraphics[width=0.98\linewidth]{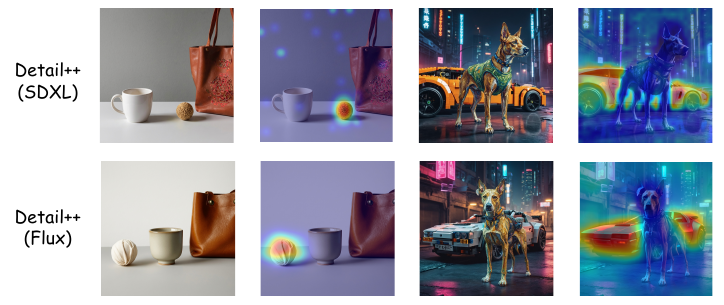}
  \caption{\rv{Visualization of adding the texture attribute ``silk'' to the small subject ``ball'' and adding the style attribute ``Lego-style'' to the overlapping subject ``sports car''.}}
  \label{app: fig case study}
\end{figure}

\label{app: case study}
\subsection{Case Study}
\noindent\rv{\textbf{Critical Cases.} In this section, we present Detail++ results on corner cases involving small subjects and overlapping subjects. As shown in Fig.\ref{app: fig case study}. We observe that the model can accurately localize the target subject to be modified during generation, enabling Detail++ to remain effective in these challenging scenarios.}

\begin{figure}[]
  \centering
  \includegraphics[width=\linewidth]{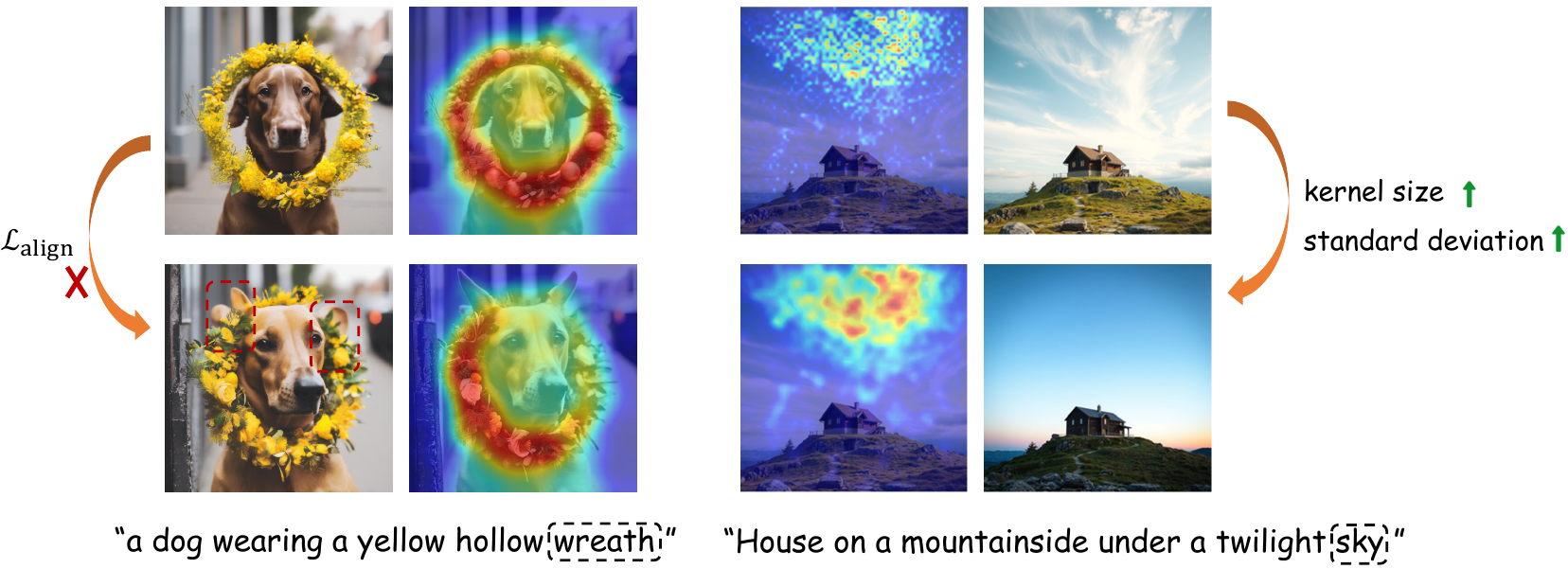}
  \caption{\rvtwo{Failure cases and mitigations for Detail++.
  \textbf{Left:} Detail++ (SDXL) on ``a dog wearing a yellow hollow wreath.''
  The centroid alignment loss $L_{\text{align}}$ shifts attention toward the hollow interior of the wreath (bottom). Consequently, this may lead to subject blending as highlighted by the red bounding boxes.
  \textbf{Right:} Detail++ (Flux.1) on ``House on a mountainside under a twilight sky.''
  Salt-and-pepper noise in background attention maps (top) degrades mask extraction.
  Stronger Gaussian blur during mask extraction suppresses the noise and yields cleaner masks (bottom).}}
  \label{fig:failure-cases}
\end{figure}

\noindent\rvtwo{\textbf{Qualitative Failure Case Analysis.}
We present representative failure cases of Detail++ on both U-Net-based and DiT-based backbones, as illustrated in Fig.~\ref{fig:failure-cases}.
For Detail++ (SDXL), test-time optimization is generally employed to prevent subject blending.
However, the centroid alignment loss $L_{\text{align}}$ can produce suboptimal results for hollow subjects, because the geometric centroid of such objects lies in the hollow interior rather than on the object surface.
For example, for the prompt ``a dog wearing a yellow hollow wreath'', $L_{\text{align}}$ pulls the wreath's attention region toward its one point, yielding a distorted wreath rather than a coherent ring-shaped structure.
In such cases, test-time optimization should be disabled for the affected subject.
For Detail++ (Flux.1), we observe that the image-to-token attention maps for background tokens exhibit salt-and-pepper noise.
This noise reduces the accuracy of binary attention mask extraction and consequently degrades attribute injection quality.
Therefore, increasing the standard deviation or kernel size of the Gaussian blur applied during mask extraction effectively mitigates this issue.}

\subsection{Combination with Other Methods}\label{COmbination}
Our method can also be successfully integrated with other commonly used techniques in AIGC, such as Control-net~\cite{zhang2023adding}. In this section, we demonstrate the efficiency of our approach when combined with Control-net, as shown in Fig.~\ref{app: with-control-net}.

\begin{table*}[t]
  \centering
  \small
  \setlength{\tabcolsep}{6pt}
  \renewcommand{\arraystretch}{1.18}
  \caption{\rv{Prompt pool used for image generation in the questionnaire. Attribute Binding prompts are short, natural generation prompts with salient subject-attribute bindings. Style Binding prompts follow the three templates and candidate sets (styles/subjects/backgrounds) in Tab.\ref{app:benchmark construction}. Image Quality prompts contain multiple subjects with explicit attributes, enriched context, and aesthetic/photographic cues for high-fidelity generation.}}
  \label{tab:prompt_pool}
  \begin{tabular}{p{0.19\linewidth} p{0.78\linewidth}}
    \toprule
    \textbf{Dimension} & \textbf{Prompts (12 per dimension)} \\
    \midrule
    
    \textbf{Attribute Binding} &
    \textbf{AB-01} A \textbf{ginger cat} wearing a \textbf{blue collar} sits beside a \textbf{black dog} wearing a \textbf{red bandana} on a wooden floor. \newline
    \textbf{AB-02} A \textbf{robot} with \textbf{round eyes} holds a \textbf{yellow balloon}, while an \textbf{astronaut} with a \textbf{square visor} holds a \textbf{red flag}. \newline
    \textbf{AB-03} A \textbf{woman} in a \textbf{green coat} holds a \textbf{white umbrella}, and a \textbf{man} in a \textbf{black coat} carries a \textbf{brown suitcase}. \newline
    \textbf{AB-04} On a table, a \textbf{red mug} sits next to a \textbf{yellow teapot}, with a \textbf{silver spoon} placed in front. \newline
    \textbf{AB-05} A \textbf{blue car} is parked in front of a \textbf{yellow train}, with a small \textbf{red traffic cone} between them. \newline
    \textbf{AB-06} A \textbf{panda} wearing a \textbf{striped scarf} stands next to a \textbf{tiger} wearing a \textbf{polka-dot scarf} by a wooden bench. \newline
    \textbf{AB-07} A \textbf{white vase} holds a \textbf{red rose} on the left and a \textbf{purple tulip} on the right, on a marble tabletop. \newline
  \textbf{AB-08} A glass bowl with three fruits: a \textbf{green apple}, a \textbf{yellow banana} in the center, and a \textbf{purple grape}. \newline
  \textbf{AB-09} A \textbf{black piano} with a \textbf{white sheet of music}, and a \textbf{brown violin} resting on a \textbf{red chair}. \newline
  \textbf{AB-10} A \textbf{silver spaceship} hovers above a \textbf{gray building} with a \textbf{blue door} under an \textbf{orange sunset} sky. \newline
  \textbf{AB-11} A \textbf{brown horse} with a \textbf{red saddle} stands beside a \textbf{white unicorn} with a \textbf{blue saddle} in a grassy field. \newline
  \textbf{AB-12} An \textbf{orange robot} stands on the left, and a \textbf{teal bicycle} stands on the right, under a \textbf{purple streetlight} at night. \\
  
  \midrule
  
  \textbf{Style Binding} &
  \textbf{SB-01} A \textbf{watercolor} style \textbf{cat} in a \textbf{cyberpunk} style \textbf{city} background. \newline
  \textbf{SB-02} An \textbf{ukiyo-e} style \textbf{horse} in an \textbf{ink painting} style \textbf{forest} background. \newline
  \textbf{SB-03} A \textbf{pixel art} style \textbf{robot} in an \textbf{isometric} style \textbf{factory} background. \newline
  \textbf{SB-04} A \textbf{chalk} style \textbf{spaceship} in a \textbf{charcoal drawing} style \textbf{space} background. \newline
  \textbf{SB-05} A \textbf{Lego} style \textbf{astronaut} next to a \textbf{graffiti} style \textbf{bicycle}. \newline
  \textbf{SB-06} An \textbf{anime} style \textbf{woman} next to a \textbf{low-poly} style \textbf{car}. \newline
  \textbf{SB-07} An \textbf{oil painting} style \textbf{man} next to a \textbf{sketch} style \textbf{piano}. \newline
  \textbf{SB-08} An \textbf{ink painting} style \textbf{panda} next to a \textbf{watercolor} style \textbf{train}. \newline
  \textbf{SB-09} A \textbf{cyberpunk} style \textbf{dog} and a \textbf{watercolor} style \textbf{cat} in an \textbf{isometric} style \textbf{city} background. \newline
  \textbf{SB-10} A \textbf{pixel art} style \textbf{robot} and a \textbf{Lego} style \textbf{astronaut} in a \textbf{graffiti} style \textbf{marketplace} background. \newline
  \textbf{SB-11} An \textbf{anime} style \textbf{tiger} and an \textbf{ukiyo-e} style \textbf{unicorn} in an \textbf{ink painting} style \textbf{forest} background. \newline
  \textbf{SB-12} An \textbf{oil painting} style \textbf{horse} and a \textbf{sketch} style \textbf{bicycle} in a \textbf{low-poly} style \textbf{desert} background. \\
  
  \midrule
  
  \textbf{Image Quality} &
  \textbf{IQ-01} Cinematic studio portrait: a \textbf{woman in an emerald dress} holding a \textbf{silver camera}, beside a \textbf{man in a navy suit} with a \textbf{red tie}; soft key light, gentle rim light, 85mm look, clean bokeh, natural skin texture. \newline
  \textbf{IQ-02} Golden-hour street portrait: an \textbf{elderly man with round glasses} wearing a \textbf{blue scarf} holds a \textbf{brown leather book}; warm backlight, shallow DOF, film-like color, no halos, crisp edges on glasses. \newline
  \textbf{IQ-03} Premium product still life on white marble: a \textbf{matte black wristwatch}, a \textbf{gold ring}, and a \textbf{clear glass perfume bottle}; softbox lighting, accurate reflections/refraction, sharp micro-details, clean shadows. \newline
  \textbf{IQ-04} Cozy cafe scene by window light: a \textbf{latte with leaf art} next to a \textbf{red notebook} and a \textbf{black pen}; shallow DOF, natural highlights, realistic foam texture, tidy composition (rule of thirds). \newline
  \textbf{IQ-05} Rainy neon night street: a \textbf{yellow taxi} passing a \textbf{person with a blue umbrella}, with \textbf{pink neon} reflecting on wet asphalt; realistic glare control, sharp signage text, no banding in shadows. \newline
  \textbf{IQ-06} Modern kitchen interior: a \textbf{stainless kettle} beside a \textbf{white mug} and a \textbf{green plant}; wide-angle but straight lines, realistic global illumination, clean materials, no warped geometry. \newline
  \textbf{IQ-07} Editorial food photo: a \textbf{ramen bowl with steam} next to \textbf{wooden chopsticks} and a \textbf{blue side dish}; glossy broth highlights, realistic steam translucency, crisp ceramic edges, balanced color. \newline
  \textbf{IQ-08} Ultra-detailed macro: a \textbf{red ladybug with black spots} on a \textbf{green leaf} with \textbf{clear dew drops}; razor-sharp focal plane, smooth bokeh, realistic micro-texture, no oversharpening. \newline
  \textbf{IQ-09} Wildlife shot: a \textbf{turquoise-and-yellow parrot} perched on a \textbf{brown branch} with \textbf{green leaves} behind; sharp eye catchlight, feather micro-details, natural saturation, clean background bokeh. \newline
  \textbf{IQ-10} Motion photography: a \textbf{cyclist in a red jersey} wearing a \textbf{white helmet} rides past \textbf{orange streetlights}; subject sharp, controlled background motion blur, no duplicated limbs, realistic shutter effect. \newline
  \textbf{IQ-11} Landscape at dusk: a \textbf{wooden cabin} beside a \textbf{mirror-like lake} with an \textbf{orange canoe} near the shore; crisp treeline, subtle ripples, atmospheric perspective, HDR without artifacts. \newline
  \textbf{IQ-12} Long-exposure night scene: a \textbf{small cabin with warm window glow} and a \textbf{silver telescope} under the Milky Way; realistic stars, low noise, sharp roof edges, smooth highlight roll-off. \\
  
  \bottomrule
\end{tabular}
\end{table*}

\clearpage
\section{Example of LLM Decomposing Prompts}
\label{llm-decomposition temmplate}
\rv{Here, we provide an example of prompt used in LLM decomposing, this is in Config.B formation we discussed in Sec.~\ref{app:sps management}, and is also the formation we used in test.}
\begin{Verbatim}[fontsize=\footnotesize, breaklines=true]
  """**Detailed Instruction Prompt for Decomposing Image Descriptions**
  You are provided with an original prompt that describes an image containing one or more subjects with detailed attributes (such as colors, clothing, objects, etc.). Your task is to generate a series of sub-prompts that decompose the original prompt into simpler, attribute-focused branches. Follow the steps and rules below exactly:
  1. **Output Format Requirements:**
  - **First Line:** 
 - Begin with `[original]` followed by a space and then the complete original prompt exactly as provided.
 - **Subsequent Lines:** 
 - Each additional line must start with `[sub-index][subject]` where:
 - `sub-index` is a sequential number starting from 0.
 - `subject` is a keyword that indicates which subject's detailed attribute is being highlighted. If the attribute added is global, like background, use `None`. For the first branch, use `None` as the subject keyword.
 - **Line Separation:** 
 - Each sub-prompt must appear on its own line.
2. **Decomposition Rules:**
 - **Generic Version ([sub-0][None]):** 
 - Create a version of the prompt that has all specific detailed attributes (e.g., color adjectives, style adjectives) removed. This produces a simplified, generic description of the scene.
 - **Attribute-Specific Branches:** 
 - For every distinct subject in the original prompt that has a specific attribute, generate a branch that reintroduces that particular attribute while keeping all other subjects in their generic state.
 - Each branch must re-add the attribute detail for only one subject. For example, if the original prompt mentions a “red hat” on one subject and a “blue tracksuit” on another, then:
 - One branch should reintroduce “red” for the hat.
 - Another branch should reintroduce “blue” for the tracksuit.
 - The keyword inside the brackets (after the sub-index) should indicate the subject whose attribute is restored (e.g., `hat`, `tracksuit`, `car`, etc.).
3. **General Guidelines:**
 - **Consistency:** 
 - Ensure that the modified sub-prompts are logically consistent with the original description. Only one attribute should be reintroduced per branch, while all other attribute details remain generic.
 - **Precision:** 
 - Follow the exact fixed format with square brackets and no extra characters or commentary.
 - **No Extra Text:** 
 - Do not include any explanations, notes, or additional commentary in the output. The final output should only contain the sub-prompts as specified.
 - **Output format:** 
 - The output should be a JSON object with a single key `variants` that contains a list of sub-prompts. 
4. **Example to Follow:**
Given the original prompt: 
```
a man wearing a red hat and blue tracksuit is standing in front of a green sports car
```
The output should be:
```
{"variants":
    [
        [original] a man wearing a red hat and blue tracksuit is standing in front of a green sports car
        [sub-0][None] a man wearing a hat and tracksuit is standing in front of a sports car
        [sub-1][hat] a man wearing a red hat and tracksuit is standing in front of a sports car
        [sub-2][tracksuit] a man wearing a hat and blue tracksuit is standing in front of a sports car
        [sub-3][car] a man wearing a hat and tracksuit is standing in front of a green sports car
    ]
}
```
5. **Another Example to Follow:**
Given the original prompt: 
```
In a cyberpunk style city night, a VanGogh-style hound dog is standing in front of a lego-style sports car
```
The output should be:
```
{"variants":
    [
        [original] In a cyberpunk style city night, a VanGogh-style hound dog is standing in front of a Lego-style sports car
        [sub-0][None] In a city night, a hound dog is standing in front of a sports car
        [sub-1][None] In a cyberpunk style city night, a hound dog is standing in front of a sports car
        [sub-2][hound dog] In a city night, a VanGogh-style hound dog is standing in front of a sports car
        [sub-3][car] In a city night, a hound dog is standing in front of a Lego-style sports car
    ]
}
```
6. **Task Summary:**
 - Your task is to read the given original prompt and output a set of sub-prompts using the format above.
 - The first sub-prompt ([sub-0][None]) should be the fully generic version.
 - Each subsequent sub-prompt should selectively reintroduce one detailed attribute corresponding to a subject from the original prompt.
Now, use this detailed instruction prompt to generate the decomposed sub-prompts for any provided original image description.
---
"""
\end{Verbatim}